\documentclass{article}

\usepackage{arxiv}

\usepackage[utf8]{inputenc} 
\usepackage[T1]{fontenc}    
\usepackage{hyperref}       
\usepackage{url}            
\usepackage{booktabs}       
\usepackage{diagbox}

\usepackage{amsfonts}       
\usepackage{nicefrac}       
\usepackage{microtype}      
\usepackage{amsmath, amssymb}
\usepackage{cleveref}       
\usepackage{lipsum}         
\usepackage{graphicx}
\usepackage[style=numeric,sorting=none,giveninits=true,maxnames=99]{biblatex}
\usepackage{doi}
\usepackage{float}
\usepackage{subfigure}
\usepackage{tikz, pgfplots}
\usetikzlibrary{positioning}
\usepackage{diagbox}    
\pgfplotsset{compat=1.18}
\pgfplotsset{
  targetstyle/.style={
    thick,
    opacity=0.7
  }
}
\pgfplotsset{
  estimatestyle/.style={
    thick,
    mark options={solid},
    only marks,
  }
}
\pgfplotsset{
    legend style={
    /tikz/every even column/.append style={
        column sep=.2cm,
    }
    },
    legend image code/.code={
        \draw[mark repeat=2,mark phase=2]
        plot coordinates {
        (0cm,0cm)
        (0.15cm,0cm)        
        (0.3cm,0cm)         
        };%
    }
}
\pgfplotsset{
  every axis plot/.append style={
    mark size=1.5pt  
  }
}

\newcommand{\diff}{\text{d}}            
\newcommand{\R}{\mathbf{R}}             
\newcommand{\mat}[1]{\boldsymbol{#1}}   

\title{Physics-Informed Regression: Parameter Estimation in Parameter-Linear Nonlinear Dynamic Models}

\newif\ifuniqueAffiliation
\uniqueAffiliationtrue

\ifuniqueAffiliation 
\author{
    \href{https://orcid.org/0009-0000-6295-6428}{\includegraphics[scale=0.06]{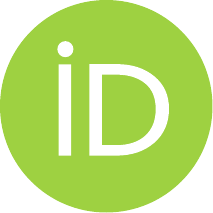}\hspace{1mm}Jonas Søeborg Nielsen} \\
    \href{mailto:jonassoeborgnielsen@icloud.com}{jonassoeborgnielsen@icloud.com}    
    \And
    \href{https://orcid.org/0009-0001-0489-0325}{\includegraphics[scale=0.06]{orcid.pdf}\hspace{1mm}Marcus Galea Jacobsen} \\
    \href{mailto:marcusgaleajacobsen@gmail.com}{marcusgaleajacobsen@gmail.com}
    \And
    Albert Brincker Olson \\
    \href{mailto:alboarddk@gmail.com}{alboarddk@gmail.com}
    \And
    Mads Peter Sørensen \\
    \href{mailto:mpso@dtu.dk}{mpso@dtu.dk}
    \And
    \href{https://orcid.org/0000-0001-8626-1575}{\includegraphics[scale=0.06]{orcid.pdf}\hspace{1mm}Allan Peter Engsig-Karup} \\
    \href{mailto:apek@dtu.dk}{apek@dtu.dk}
}
\else
\usepackage{authblk}

\newbox{\orcid}\sbox{\orcid}{\includegraphics[scale=0.06]{orcid.pdf}} 
\author[1]{%
	\href{https://orcid.org/0000-0000-0000-0000}{\usebox{\orcid}\hspace{1mm}David S.~Hippocampus\thanks{\texttt{hippo@cs.cranberry-lemon.edu}}}%
}
\author[1,2]{%
	\href{https://orcid.org/0000-0000-0000-0000}{\usebox{\orcid}\hspace{1mm}Elias D.~Striatum\thanks{\texttt{stariate@ee.mount-sheikh.edu}}}%
}
\affil[1]{Department of Computer Science, Cranberry-Lemon University, Pittsburgh, PA 15213}
\affil[2]{Department of Electrical Engineering, Mount-Sheikh University, Santa Narimana, Levand}
\fi

\renewcommand{\shorttitle}{Physics-Informed Regression}

\hypersetup{
pdftitle={Physics-Informed Regression},
pdfauthor={Jonas S.~Nielsen},
pdfkeywords={PIR, SIR, Parameter estimation},
}

\begin{document}
\maketitle
\begin{abstract}
     We present a new efficient hybrid parameter estimation method based on the idea, that if nonlinear dynamic models are stated in terms of a system of equations that is linear in terms of the parameters, then regularized ordinary least squares can be used to estimate these parameters from time series data. We introduce the term {\em physics-informed regression} (PIR) to describe the proposed data-driven hybrid technique as a way to bridge theory and data by use of ordinary least squares to efficiently perform parameter estimation of the model coefficients of different parameter-linear models; providing examples of models based on nonlinear ordinary equations (ODE) and partial differential equations (PDE).
     The PIR method is used for parameter estimation and tested and compared against the related technique, physics-informed neural networks (PINN), an alternative hybrid method also based on incorporating an assumed model, and possibly utilizing available data, into a loss function to perform parameter estimation and determine a neural network that approximates the solution of the governing equations. The focus is on parameter estimation on a selection of ODE and PDE models, each illustrating performance in different model characteristics. For two relevant epidemic models of different complexity and number of parameters, PIR is tested and compared against PINN, both on synthetic data generated from known target parameters and on real public Danish time series data collected during the COVID-19 pandemic in Denmark. Both methods were able to estimate the target parameters, while PIR showed to perform noticeably better, especially on a compartment model with higher complexity, such as a S3I3R model. Given the difference in computational speed, it is concluded that the PIR method is superior to PINN for the models considered. It is also demonstrated how PIR can be applied to estimate the time-varying parameters of a compartment model that is fitted using real Danish data from the COVID-19 pandemic obtained during a period from 2020 to 2021. The study shows how data-driven and physics-informed techniques may support reliable and fast -- possibly real-time -- parameter estimation in parameter-linear nonlinear dynamic models.
\end{abstract}



\section{Introduction}

Dynamical systems, governed by differential equations that may be nonlinear, are central in a wide range of science and engineering fields, including economics, biology, physics, and epidemiology. For tasks such as forecasting and interpretability of the underlying processes, it is useful to fit the model parameters to the experimental data, which may be a challenging task, especially in cases where the models are nonlinear, the data contain noise, or if the models contain many parameters \cite{Transtrum2009WhyChallenging}. In many applications, the primary objective is often to forecast the progression of a system by solving the initial value problem (IVP) defined by governing ODEs or PDEs. However, the model parameters are typically either partially or fully unknown and must be inferred from observed data. In this work, given the availability of such data, the focus is on the inverse problem of estimating the underlying parameters in well-known dynamical systems. These include the ODEs of the Lotka-Volterra predator-prey system and the spread-of-disease SIR model, as well as the PDE system of Naviers-Stokes equations for incompressible fluid flow.

Estimation of model parameters in differential equations is a common problem, and there are several solution methods  \cite{Ramsay2007ParameterApproach}. Most of these methods are based on iterative nonlinear optimization techniques, e.g., formulating the problem as a boundary value problem and then use ODE solvers as a means for single-shooting \cite{Isaacson1994AnalysisMathematics} or multi-shooting techniques for nonlinear ODEs \cite{Bock1984AProblems,Peifer2007ParameterShooting,Bock2015DirectModels}, or likelihood-based strategies~\cite{Kato2020Likelihood-basedModel, Liu1989OnOptimization, Beck1977ParameterScience, Bard1974NonlinearEstimation, Hestenes1969MultiplierMethods, Metropolis1953EquationMachines, Elerian2001LikelihoodDiffusions}. While these approaches can yield accurate parameter estimates for a broad range of problems, they are iterative, converging towards optimality through incremental steps, thereby increasing the computational load. 
In addition, for systems with chaotic dynamics, it is common for strategies that rely on solving initial value problems numerically to experience difficulties with accurate estimation due to a lack of precision \cite{Baake1992FittingData,Rackauckas2025HowLifestyle}.

In the field of data science, approaches using neural networks have proven useful for solving forward and inverse problems, e.g., see some of the original works~\cite{Lee1990NeuralEquations,
Psichogios1992AModeling,
Meade1994SolutionNetworks,Meade1994TheNetworks, 
Dissanayake1994NeuralnetworkbasedEquations,Lagaris1997ArtificialEquations,
Rudd2013SolvingNetworks,
Sirignano2018DGM:Equations} and further analyzed and developed for numerous applications over the past decades. Two such approaches were recently formalized in the form of neural ordinary differential equations (NODE)~\cite{Chen2018NeuralEquations} and the physics-informed neural network (PINN)~\cite{Raissi2019Physics-informedEquations} framework. Unlike more traditional methods, PINNs make it possible to leverage both observed data and assumed models defined in terms of an ODE or PDE system to simultaneously estimate parameters and a surrogate solution. Furthermore, in recent years, PINNs have been adopted in science and engineering, e.g., for nonlinear systems of ODEs such as infectious disease modeling \cite{Shaier2022Data-DrivenNetworks}, and for nonlinear systems of PDEs such as acoustics \cite{Borrel-Jensen2023Physics-informedBoundaries}, spread of wildfires \cite{Vogiatzoglou2025Physics-informedSpreading}, etc. The use of techniques in which learning algorithms are limited to satisfy physical laws belongs to the growing field of physics-informed learning, cf. review \cite{Karniadakis2021Physics-informedLearning}.

In recent years, the frameworks of sparse identification of nonlinear dynamics (SINDy) \cite{Brunton2016DiscoveringSystems,Kaheman2020SINDy-PI:SINDy-PI} and partial differential equation functional identification of nonlinear dynamics (PDE-find)~\cite{Rudy2016Data-drivenEquations}, have gained strong traction in scientific applications and are widely regarded as state-of-the-art methods for discovering nonlinear dynamics from time series data. Assuming an underlying parameter-linear model of governing differential equations, both frameworks utilize a user-defined library of basis functions and combine model regression with parameter pruning to identify a set of sparse differential equations that best describe the observed data. Despite inherently performing parameter estimation, SINDy and PDE-find are predominantly used for model discovery, as they rely exclusively on data and do not assume an {\em a priori} known model structure.

\subsection{Contributions}
We propose a new closed-form parameter estimation method for a specific subclass of ODE systems described in Section \ref{sec:systemsofeqs} characterized by a parameter-linear structure. We take advantage of that the class of parameter-linear models is easy to optimize because techniques like linear regression or least squares apply directly. By posing the parameter estimation problem starting from the models it is possible to incorporate inductive bias ({\em domain knowledge}) to make the regression problem {\em physics-informed}. Similarly, to SINDy, many dynamical systems may comprise nonlinear ODEs, and can often be formulated as a matrix-vector product between a state matrix and a vector of parameters. Provided that the states and their respective time derivatives can be adequately estimated from available time series data, the model parameters can therefore be estimated directly using the closed-form solution of ordinary least squares (OLS)~\cite{Kiers1997WeightedAlgorithms} regression. Compared to the aforementioned methods, this approach provides a significantly faster and robust alternative for data-driven parameter estimation. Conceptually, this approach resembles a SINDy model using a function library that contains only the known terms of the governing ODE system. However, it is still possible to include unknown candidate terms in each individual equation and infer their coefficients.

Drawing inspiration from the recent PINN framework \cite{Raissi2019Physics-informedEquations} briefly introduced in Section \ref{sec:PINNs}, this method is named {\em physics-informed regression} (PIR) and is detailed in Section \ref{sec: PIR}. We evaluated our proposed method for parameter estimation compared to PINNs, which have seen significant adoption in recent years in computational science and engineering. In this context, the term {\em physics-informed} denotes the assumption of an underlying dynamic system that can be modeled, where the differential equations governing the model are integrated into the residual error function used for model evaluation. In such models, the model parameters may be partially or fully unknown, motivating the utilization of available time series data for parameter estimation.

This study focuses on using observations to estimate parameters within a presumed underlying model and, therefore, is generally not concerned with the validity of the assumed dynamics via specific choices of models assumed to adequately describe the observations. The experiments are carried out on two epidemic models described in Section \ref{sec:modalsanddata} that contain two and seven parameters, respectively. In Sections \ref{sec:PIRexpODE} and \ref{sec:PIRexpPDE} we report on the parameter estimation results obtained using the PIR method, and in Section \ref{sec:PINN} we use the PINN approach.

\section{Systems of Nonlinear Differential Equations}
\label{sec:systemsofeqs}
Systems of differential equations are a fundamental tool in mathematical modeling to describe dynamical processes. While linear systems yield elegant solutions, real-world dynamics often exhibits nonlinear behavior, where the relationships between variables are not simply proportional. Time-dependent ODE systems consisting of $m$ ODEs are formulated in terms of the time derivative $\frac{\diff {\bf x}}{\diff t}$ of a state vector ${\bf x}(t)\in\mathbb{R}^n$ with $n$ states and $k$ model parameters ${\mat\omega}\in{\bf R}^k$, formally written in the form of the general initial value problem (IVP) 
\begin{align}
\label{eq:IVP_general}
    \frac{\diff {\bf x}}{\diff t} = {\bf f}({\bf x}(t); t, {\mat\omega}), \quad {\bf x}(0)={\bf x}_0, \quad t\geq0,
\end{align}
where it is assumed that ${\bf f}:C^1(\mathbb{{R}^+}, \mathbb{R}^n)\to C^0(\mathbb{{R}^+}, \mathbb{R}^m)$\footnote{Map from the set of real valued $n$-dimensional differentiable vector functions to the set of real valued $m$-dimensional continuous vector functions.} is Lipschitz continuous with respect to ${\bf x}$.

To apply the proposed physics-informed regression method, it is assumed that the ODE system that best describes the data is assumed to be parameter-linear. Specifically, the function ${\bf f}$ is formulated as the matrix-vector product \begin{align}
\label{eq:IVP_linear}
    \frac{\diff {\bf x}}{\diff t} = A({\bf x}(t);t) {\mat\omega}, \quad {\bf x}(0)={\bf x}_0, \quad t\geq0,
\end{align}
where the system matrix $A({\bf x}(t);t)$ consists of basis functions independent of ${\mat\omega}$. For example, common spread-of-disease compartment models can be expressed as in \eqref{eq:IVP_linear} with typical examples given in terms of the so-called SIR models, cf. Section \ref{sec:modalsanddata}.

Across science and engineering, there are many processes that can be modeled by a system of equations described by ordinary or partial differential equations (PDEs). To bridge PDEs with the framework of ODEs, one may collect snapshots of solutions wrt. time that can be described by PDEs. A common approach involves extracting time series data at fixed spatial sensor locations ${\bf x}_i(t)$, $i=1,...,N_s$, thereby reducing the PDE to a system of ODEs in time. This reduction enables the application of ODE-based regression methods to estimate the underlying dynamics from the observed data. A well-known example of this idea is the method of lines (MoL) often used for the numerical discretization of time-dependent PDEs, in which the spatial domain is first discretized into a finite set of grid points. At each of these points, the spatial derivatives are approximated, e.g. via finite differences, transforming the PDE into a high-dimensional system of ODEs that govern the temporal evolution at the discrete spatial locations. This approach allows for a unified treatment of spatio-temporal systems, where local gradients and time-dependent behavior can be captured within a physics-informed differential equation framework that incorporates models stated as IVPs \eqref{eq:IVP_general}. 

\section{Physics-Informed Regression (PIR) using Ordinary Least Squares} \label{sec: PIR}

In the following, we introduce the physics-informed regression (PIR) method based on ordinary least squares.
The goal of PIR is to learn the functional form $f(\cdot)$ in \eqref{eq:IVP_general} by assuming a parameter-linear model as in \eqref{eq:IVP_linear} from time series data.

Given the $i$'th state ${\bf x}_i={\bf x}(t_i)$ and its temporal derivative $\frac{\diff {\bf x}_i}{\diff t}=\frac{\diff {\bf x}}{\diff t}|_{{\bf x}={\bf x}_i}$, the ODE residuals for a given set of parameters $\omega$ are defined by \eqref{eq:ODE_residual} as the difference between the observed derivative and the predicted derivative of the model, calculated by equation \eqref{eq:IVP_linear},
\begin{equation} \label{eq:ODE_residual}
    {\bf r}_i({\mat\omega}_i) = A_{i} {\mat\omega}_i-\frac{\diff {\bf x}_i}{\diff t},
\end{equation}
where $A_i=A({\bf x}_i)$ is the coefficient matrix computed using state ${\bf x}_i$. Assuming that state ${\bf x}_i$ and its neighboring states can be sampled sufficiently smooth, the temporal derivative $\frac{d{\bf x}_i}{dt}$ can be reasonably approximated by the second-order accurate central finite difference method given by \eqref{eq:central_difference},
\begin{equation}\label{eq:central_difference}
    \frac{\diff {\bf x}_i}{\diff t} \approx \frac{{\bf x}_{i+1}-{\bf x}_{i-1}}{t_{i+1}-t_{i-1}}.
\end{equation}
For time series data considered in this work, $\Delta t=t_{i+1}-t_{i-1}$ is the time interval between measurements ${\bf x}_{i+1}$ and ${\bf x}_{i-1}$. Remark that it would be possible to use irregular time measurements for the approximation of temporal derivatives, although it is not considered in this work. In an ideal noise-free setting, the approximation error of the central difference method decreases as $\Delta t$ decreases. 

However, this might not be the case in practice. Furthermore, if the data are polluted by noise, the error will be passed on to the approximation \eqref{eq:central_difference}. The approach to mitigate this issue is to combine data smoothing~\cite{Lodhi2022NumericalMethod} with collocation methods~\cite{Moreno-Martin2024CollocationSystems} or use techniques such as gradient matching~\cite{Ellner2002FittingMatching} to obtain more robust approximations of derivatives. In early work, it was also proposed to fit splines to the data and after this the data were fit numerically to estimate the parameters of differential equations via a least square method \cite{Varah1982AEquations}.

In the context of PDEs, relying on finite difference methods has some sensitivity to noise in the data. However, more data may sometimes support the reduction of the impact of noise when using regression techniques \cite{Raissi2018DeepEquations}. In related work on data-driven discovery of equations from data, it is reported that denoising snapshot data could be done via SVD \cite{Rudy2019Data-drivenEquations}. However, such denoising or noise-filtering strategies are not considered in this work.



The goal is to estimate each of the model parameters ${\mat\omega}_i$ so that the squared error $||{\bf r}_i({\mat\omega}_i)||_2^2$ is minimized. As linear least-squares problems are convex, any extrema of the squared error function are a global minimizer. As the model should be fitted to observed data that yield unique parameters, the system is required not to be under-determined, which would result in infinitely many solutions. Thus, $A_i$ should be tall and of full rank, that is, $k\ll m$ and $\text{rank}(A_i)=k$. According to Theorem \ref{sec:regLSQproof}, the unique global minimizer ${\mat\omega}_i^*$ is given by \cite{Cheney2020NumericalComputing} 
\begin{equation}\label{eq: OLS}
    {\mat\omega}_i^* = \underset{{\mat\omega}_i \in \Omega}{\text{argmin}} ||{\bf r}_i({\mat\omega}_i)||_2^2 = (A_i^T A_i)^{-1}A_i^T \frac{\diff {\bf x}_i}{\diff t}.
\end{equation}
Typically, $A_i$ is considered tall when there are more compartments than parameters, which is not commonly the case in epidemic modeling, since each compartment usually introduces at least one additional parameter. To avoid the system being under-determined, information from adjacent states is incorporated by linking temporal state information across discrete time points. Formally, this is achieved by vertically concatenating them as given by 
\begin{equation}\label{eq:overdetermined_system}
    {\mat A}_I = \begin{bmatrix}
        A_{i}\\
        A_{i+1} \\
        \dots \\
        A_{j}
    \end{bmatrix}, \quad
    {\mat b}_I = \begin{bmatrix}
        \scriptstyle\frac{\diff {\bf x}_{i}}{\diff t}\\
        \scriptstyle\frac{\diff {\bf x}_{i+1}}{\diff t}\\ 
        \dots \\ 
        \scriptstyle\frac{\diff {\bf x}_{j}}{\diff t}
    \end{bmatrix}, \quad
    {\mat r}_I({\bf\omega}_I) = \begin{bmatrix}
        r_i({\bf\omega}_I)\\
        r_{i+1}({\bf\omega}_I) \\
        \dots \\
        r_j({\bf\omega}_I)
    \end{bmatrix}, \quad
\end{equation}
with $i<j$ and $I = \{i, i+1,\dots,j \}$ denoting a discrete set of consecutive time points. Additionally, this extension immediately enables fitting models to multiple states according to 
\begin{equation}\label{eq:OLS_multi}
    \omega_I^* = (A_I^T A_I)^{-1}A_I^T {\bf b}_I.
\end{equation}

Before model fitting, one or more model parameters may already be known apriori and should therefore not be estimated. To accommodate this, ${\mat\omega}_I$ and $A_I$ are partitioned into a known and unknown part, as given by \eqref{eq:known_unknown_split},
\begin{equation}\label{eq:known_unknown_split}
    {\mat\omega}_I = \begin{bmatrix}
        {\mat\omega}_{k,I} \\ {\mat\omega}_{u,I}
    \end{bmatrix}, \quad
    A_I = \begin{bmatrix}
        A_{k,I} & A_{u,I}
    \end{bmatrix}.
\end{equation}

Subsequently, the residual can be split into two
\begin{equation} \label{eq:ODE_split_residual}
    {\bf r}_I({\mat\omega}_I) = A_{k,I} {\mat\omega}_{k,I} + A_{u,I} {\mat\omega}_{u,I} - {\bf b}_I.
\end{equation}

Applying the same reasoning used to derive \eqref{eq:OLS_multi}, the corresponding expression \eqref{eq:OLS_split} can be obtained to estimate unknown parameters
\begin{equation}\label{eq:OLS_split}
    {\mat\omega}_{u,I}^* = \underset{{\mat\omega}_I \in \Omega}{\text{argmin}} ||{\bf r}_I({\mat\omega}_I)||_2^2 = (A_{u,I}^T A_{u,I})^{-1}A_{u,I}^T \underbrace{({\bf b}_I-A_{k,I} {\mat\omega}_{k,I})}_{\text{Inductive bias}}.
\end{equation}

\begin{figure}[H]
    \centering
    \includegraphics{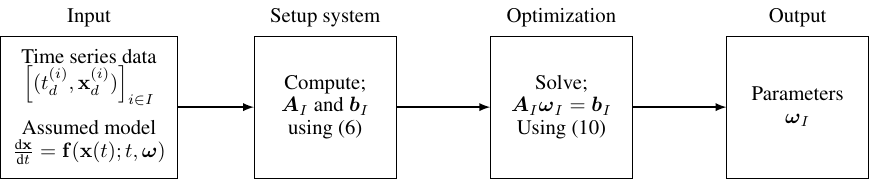}
    \caption{Graphical representation of the PIR model fitting process.}
    \label{fig:pir_training_flowchart}
\end{figure}

This constitutes the essence of the physics-informed regression (PIR) technique; with the ODE residual representing the physics-informed component as a means to incorporate domain knowledge and the ordinary least squares (OLS) minimization of residuals representing the regression component. We stress that the inductive bias here refers to the assumption that the chosen model is a reasonable way to describe the causal relationship between states that can be captured from the time series data used to perform the PIR to recover the parameter estimation of the model parameters. The PIR procedure is illustrated in Figure \ref{fig:pir_training_flowchart}.

\section{Physics-Informed Neural Networks (PINNs)}
\label{sec:PINNs}
\begin{figure}[H]
    \centering
    \includegraphics{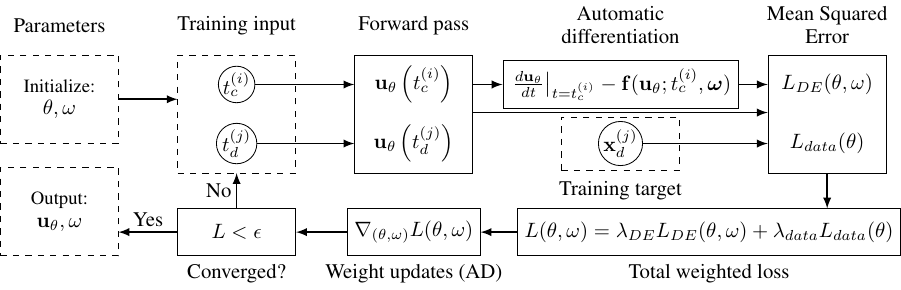}
    \caption{Graphical representation of the PINN training process. The loss terms $L_{DE}$ and $L_{data}$ are the physics and data losses respectively as defined in expressions \eqref{eq:physics_loss} and \eqref{eq:data_loss}. Start in the top left by initializing the neural network weights $\theta$ and model parameters $\omega$. Output is the trained solution surrogate network ${\bf u}_\theta$ and the estimated model parameters $\omega$.}
    \label{fig:pinn_training_flowchart}
\end{figure}

A physics-informed neural network~\cite{Raissi2019Physics-informedEquations} is an alteration of the traditional neural network training process~\cite{Rumelhart1986LearningErrors, Lecun2015DeepLearning}, designed to estimate surrogate solutions of differential equations. For time-dependent ODEs, a PINN is a map ${\bf u}_\theta: T \to {\bf R}^{d_L}$ from a temporal domain $T$ to a $d_L$-dimensional state space. In addition to the usual mean squared residual loss $L_{data}(\theta)$ given by


\begin{equation}\label{eq:data_loss}
    L_{data}(\theta) = \frac{1}{|I_{data}|}\sum_{i \in I_{data}} \|{\bf x}_i-{\bf u}_\theta(t_i)||_2^2,
\end{equation}
where $\theta$ contains the parameters of the neural network, PINNs incorporate differential equations through an additional loss term $L_{DE}$ defined by equation \eqref{eq:physics_loss}. Evaluating $L_{DE}$ only requires temporal points $t_i$, thus these can be densely sampled only at computational cost. $L_{DE}$ is a measure of how well the surrogate satisfies the ODEs, contrary to $L_{data}$ which is an absolute measure of the fit of a particular trajectory at the observed points.
\begin{equation}\label{eq:physics_loss}
    L_{DE}(\theta, {\mat\omega}) = \frac{1}{|I_{DE}|}\sum_{i \in I_{DE}} \left\|{\mat f}({\mat u}_\theta; t_i, {\mat\omega}) - \left.\frac{\diff {\mat u}_\theta}{\diff t}\right|_{t=t_i}\right\|_2^2.
\end{equation}
Composed of relative loss weights $\lambda_{DE}$ and $\lambda_{data}$, these loss terms comprise the total loss function in the expression
\begin{equation}\label{eq:pinn_loss}
    L_{PINN}(\theta, {\mat\omega}) = \lambda_{data}L_{data}(\theta) + \lambda_{DE}L_{DE}(\theta, {\mat\omega}).
\end{equation}
It is important to emphasize that PINNs employ weak enforcement of the differential equations. Thus, the network is not strictly constrained to satisfy the ODE, but instead is encouraged to do so on samples over the domain. To improve the efficiency of the multi-objective loss function \eqref{eq:pinn_loss} it can be advantageous to use weight loss balancing algorithms that update the relative weighting of the loss terms adaptively to keep the loss terms balanced in magnitude \cite{Heydari2019SoftAdapt:Functions}. The PINN training process is illustrated in Figure \ref{fig:pinn_training_flowchart}.

An advantage of using PINNs over a purely data-driven artificial neural network (ANN), is the opportunity to utilize data in combination with differential equations to estimate the parameters of an ODE model. This is due to the ODE parameters $\omega$ being updated in the same way as the ANN parameters $\theta$, using a non-linear optimization algorithm such as Adam~\cite{Kingma2014Adam:Optimization} or L-BFGS \cite{Liu1989OnOptimization}. These algorithms iteratively update the parameter values $\theta$ in a gradient-based manner, enabled by automatic differentiation (AD)~\cite{Rumelhart1986LearningErrors,GunesBaydin2015AutomaticSurvey}, to minimize the loss function $L(\theta)$. Thus, PINNs can be employed for parameter estimation in the process of approximating a solution to a set of governing ODEs.

\section{Data}
\label{sec:modalsanddata}
This work focuses primarily on parameter estimation rather than model validation; however, some models are needed to test and compare parameter estimation methods. Time-series data specific to these models are also required.

In the following sections \ref{sec:PIRexpODE} and \ref{sec:PIRexpPDE}, we present and perform experiments on a selection of models representing different properties and levels of complexity, all stated in the form of IVPs \eqref{eq:IVP_linear} with the property that the right-hand side function can be expressed as a parameter-linear model \eqref{eq:IVP_linear}. Details on the initial conditions needed to solve the specific IVPs are provided in their respective Sections.

\subsection{Time Series Data}
For the experiments, both synthetic data was produced using numerical schemes for the ODE experiments in Section \ref{sec:PIRexpODE} and Section \ref{sec:PINN}, a public data set were used for the PDE experiment in Section \ref{sec:PIRexpPDE}, and real data were used for validation of the spread-of-disease modeling. 

Synthetic data have been generated by solving IVPs from initial conditions ${\mat x}_0$ and by selection of parameters ${\mat \omega}$ (ground truth) after which temporal integration was performed using the explicit fourth-order Runge-Kutta (ERK4) method~\cite{Butcher2016NumericalEquations}, which produces solutions in different time steps, which are added to the synthetic data matrix $X$. The data matrix is chronologically ordered, that is, the row $i$ corresponds to the state ${\mat x}_i$ at time $t_i$ with $t_{i-1}<t_i<t_{i+1}$. Such synthetic data produced numerically with selected known parameters is used to verify the correctness of the algorithms by using the selected model parameters as the ground truth when evaluating the accuracy of the parameter estimation.

The real data from the COVID-19 pandemic used in Section \ref{sec:SIRrealdata} for the epidemiological models are sourced from the World Health Organization (WHO)~\cite{WorldHealthOrganization2025WHODenmark}. Since the raw data are not published in a form directly compliant with the model compartments, some pre-processing has been necessary. Details on data pre-processing are provided in Appendix \ref{sec: note on datatrans}.

\subsection{Added Noise}
To emulate the imperfections typically present in real-world data, a noise scheme is employed. Throughout the experiments, noise is added to simulated data according to
\begin{equation}\label{eq:noise}
    \text{noise} = \epsilon \cdot \max_{t\in\Omega}|\mat{x}(t)|\cdot \mathcal{N}(0,1),
\end{equation}
where $\epsilon$ is the {\em noiselevel}, $\max_{t\in\Omega}|\mat{x}(t)|$ is the magnitude of the data and $\mathcal{N}(0,1)$ is the standard normal distribution.


\section{Experiments with PIR on ordinary differential equations}
\label{sec:PIRexpODE}
Before testing PIR on real-world data, it should first test viability on synthetic data for which the exact parameters are known. For this purpose, the dynamic model itself is utilized to generate data that can be used for testing.

\subsection{Lotka-Volterra}

To demonstrate that the efficiency of the PIR method extends beyond epidemic models, we include additional well-known ODE examples from the literature. One such model is the well-known Lotka-Volterra system, which has been used in the literature to test other regression techniques~\cite{Hao2021ParameterApplication}. This model describes the population dynamics between prey and predators, denoted ${\bf x}(t)=(x(t),y(t))^T$, as follows,
\begin{align}\label{eq: Lotka-Volterra}
    \frac{\diff x}{\diff t} &= \alpha x- \beta x y, \quad
    \frac{\diff y}{\diff t} = -\gamma y + \delta x y , \quad t\geq 0,
\end{align}
where $\alpha$ and $\delta$ are parameters defining the growth/decay rates of the prey and predator populations. $\beta$ and $\gamma$ describe the effect of the interaction between the two species, i.e., predators consume prey. \eqref{eq: Lotka-Volterra} may be written in parameter-linear matrix notation as
\begin{equation}\label{eq:LV_ODE_matrix}
    \frac{\diff \mat x}{\diff t} = {\mat A}({\mat x}) {\mat\omega}, \quad
    {\mat x} = \begin{bmatrix}
    x \\
    y
    \end{bmatrix},
    \quad
    \mat{A}({\mat x}) = \begin{bmatrix}
        x & -xy & 0 & 0\\
        0 & 0 & -y & xy
    \end{bmatrix},
    \quad
    {\mat \omega} = \begin{bmatrix}
        \alpha & \beta & \gamma & \delta
    \end{bmatrix}^T, \quad t \geq 0.
\end{equation}

\begin{figure}[H]
  \centering
  \subfigure[Lotka-Volterra simulation using PIR estimated parameters from 40 equidistance points.]{%
    \includegraphics{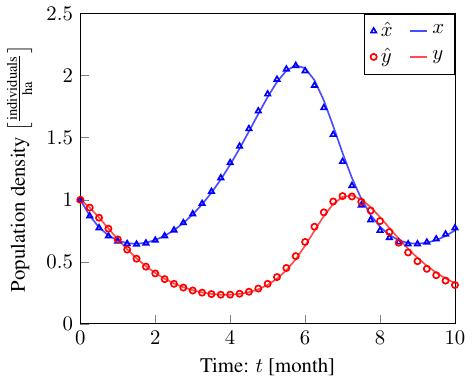}
    \label{fig:LV_simulation}
  }
  \subfigure[Parameter Convergence with increasing number of training points.]{%
    \includegraphics{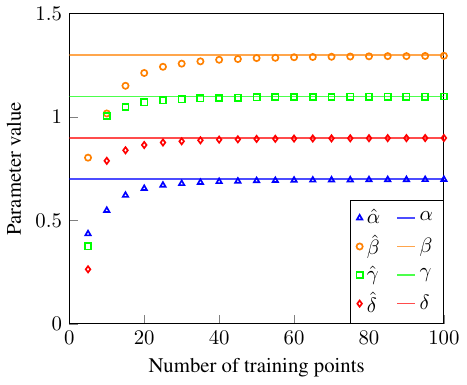}
    \label{fig:LV_parameter_convergence}
  }
  \caption{Parameter estimation of the Lotka-Volterra system using PIR.}
  \label{fig:LV_experiment}
\end{figure}

For the Lotka-Volterra simulation, a population is initialized with equal density of prey and predators.
\begin{align}
    {\mat{x}}_0 &=
    \begin{bmatrix}
        x_0 & y_0
    \end{bmatrix}^T
    = 
    \begin{bmatrix}
        1.0 & 1.0 
    \end{bmatrix}^T \frac{\text{individuals}}{\text{ha}}
\end{align}
along with the growth/decay and interaction rates
\begin{align}
    {{\mat\omega}} &= 
    \begin{bmatrix}
    \alpha & \beta & \gamma & \delta 
    \end{bmatrix}^T
    =
    \begin{bmatrix}
        0.7 & 1.3 & 1.1 & 0.9
    \end{bmatrix}^T\frac{1}{\text{month}}.\label{eq: Lotka params}
\end{align}
The IVP is solved numerically over the interval $t \in [0,10]$months for 1000 evenly spaced time points. Subsequently, PIR \eqref{eq: OLS} is used to estimate the parameters \eqref{eq: Lotka params} on a series of equispaced subsamples of data points to analyze convergence of the method. Figure \ref{fig:LV_parameter_convergence} illustrates that the parameter estimates converge to true values as the number of training points increases. Figure \ref{fig:LV_simulation} shows the result of simulating the system using parameters estimated from 40 points.

Having established parameter convergence with an increasing number of data points, we proceed to assess the impact of noise on parameter estimation. PIR is applied across various amounts of data and noiselevels, with average percent errors over 20 samples of noise reported in Table \ref{tab:lv_pir}. As expected, fewer data points and higher noise levels lead to increased estimation errors. Furthermore, the effect of noise diminishes when the number of data points is small.

\begin{figure}[H]
    \centering
    \footnotesize
    \subfigure[$\alpha$ percent error $\lbrack \% \rbrack$, true value: $\alpha=0.7\frac{1}{\text{month}}$.]{%
        \begin{tabular}{c|rrrr}
            \toprule
            \diagbox{Points}{Noise} & 0.00 & 0.01 & 0.05 & 0.10\\
            \midrule
            5 & 37.75 & 38.32 & 35.26 & 33.79\\
            10 & 21.63 & 22.18 & 22.76 & 25.62\\
            50 & 1.11 & 1.00 & 3.95 & 14.36\\
            100 & 0.29 & 0.62 & 2.38 & 7.09\\
            \bottomrule
        \end{tabular} 
        \label{tab:alpha_lv_pir}
    }
    \subfigure[$\beta$ percent error $\lbrack \% \rbrack$, true value: $\beta=1.3\frac{1}{\text{month}}$.]{%
        \begin{tabular}{c|rrrr}
            \toprule
            \diagbox{Points}{Noise} & 0.00 & 0.01 & 0.05 & 0.10\\
            \midrule
            5 & 38.12 & 38.55 & 37.43 & 35.67 \\
            10 & 21.72 & 21.94 & 23.30 & 31.57 \\
            50 & 1.15 & 1.33 & 5.36 & 18.47 \\
            100 & 0.30 & 0.83 & 4.15 & 10.67 \\
            \bottomrule
        \end{tabular} 
        \label{tab:beta_lv_pir}
    }
    \subfigure[$\gamma$ percent error $\lbrack \% \rbrack$, true value: $\gamma=1.1\frac{1}{\text{month}}$.]{%
        \begin{tabular}{c|rrrr}
            \toprule
            \diagbox{Points}{Noise} & 0.00 & 0.01 & 0.05 & 0.10\\
            \midrule
            5 & 70.77 & 71.13 & 72.02 & 78.90 \\
            10 & 12.44 & 11.71 & 15.96 & 25.94 \\
            50 & 0.67 & 1.32 & 6.38 & 19.46 \\
            100 & 0.18 & 0.44 & 6.46 & 17.73 \\
            \bottomrule
        \end{tabular} 
        \label{tab:gamma_lv_pir}
    }
    \subfigure[$\delta$ percent error $\lbrack \% \rbrack$, true value: $\delta=0.9\frac{1}{\text{month}}$.]{%
        \begin{tabular}{c|rrrr}
            \toprule
            \diagbox{Points}{Noise} & 0.00 & 0.01 & 0.05 & 0.10\\
            \midrule
            5 & 66.05 & 66.27 & 65.83 & 72.39 \\
            10 & 8.62 & 7.85 & 10.80 & 18.13 \\
            50 & 0.45 & 1.19 & 4.90 & 13.94 \\
            100 & 0.12 & 0.27 & 4.80 & 11.66 \\
            \bottomrule
        \end{tabular} 
        \label{tab:delta_lv_pir}
    }
    \caption{Percent error on PIR parameter estimates in a Lotka-Volterra model for different combinations of sample size and noiselevel. Each estimated parameter is an average of 20 samples of random noise.}
    \label{tab:lv_pir}
\end{figure}

\subsection{Epidemiological models}
The compartmental models used for this project are one class of models for spread-of-disease modeling~\cite{Mac2021ModelingModels} and are similar to those used to analyze the spread of COVID-19 in Denmark throughout 2020 reported by the Danish Statens Serums Institut~\cite{Mller2020TekniskModellerne}.

\subsubsection{SIR} \label{sec: SIR}

\begin{figure}[H]
    \centering
    \includegraphics{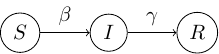}
    \caption{Flowchart of a basic SIR compartment model.}
    \label{fig:SIR_flowchart}
\end{figure}

The SIR model~\cite{Cooper2020ACommunities} is one of the most basic epidemiological models. It divides the population into three compartments, namely those \textit{susceptible} ($S$) to the disease, \textit{infected} ($I$) by the disease and finally \textit{removed} ($R$), i.e. dead or recovered. Figure \ref{fig:SIR_flowchart} illustrates the dynamical flow between compartments. The SIR model is governed by the three coupled differential equations, which are identified by two model parameters, the infection rate ($\beta$) and the recovery rate ($\gamma$) in the form
\begin{equation} \label{eq:SIR_ODE}
    \frac{\diff S}{\diff t} = -\beta\frac{SI}{N}, \quad \frac{\diff I}{\diff t} = \beta \frac{SI}{N}-\gamma I, \quad \frac{\diff R}{\diff t} = \gamma I, \quad t \geq 0.
\end{equation}
Despite \eqref{eq:SIR_ODE} being a non-linear system in terms of the state variables $S$, $I$ and $R$, it is linear in terms of the parameters $\beta$ and $\gamma$. Consequently, \eqref{eq:SIR_ODE} can be written in matrix notation as
\begin{equation}\label{eq:basic_SIR_ODE_matrix}
    \frac{\diff \mathbf{x}}{\diff t} = A({\mat x}) {\mat\omega}, \quad
    {\mat x} = \begin{bmatrix}
    S \\
    I \\
    R
    \end{bmatrix},
    \quad
    \mat{A}({\mat x}) = \begin{bmatrix}
    -\frac{SI}{N} & 0\\
    \frac{SI}{N} & -I\\
    0 & I
    \end{bmatrix},
    \quad
    {\mat \omega} = \begin{bmatrix}
    \beta\\
    \gamma
    \end{bmatrix}, \quad t \geq 0.
\end{equation}
This observation is crucial for enabling application of the physics-informed regression technique described in Section \ref{sec: PIR}.

\subsubsection{S3I3R} \label{sec: S3I3R}
\begin{figure}[H]
    \centering
    \includegraphics{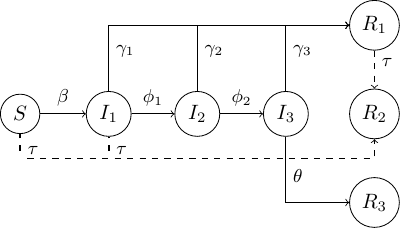}
    \caption{Flowchart of the 7 compartment S3I3R model.}
    \label{fig:S3I3R_flowchart}
\end{figure}

The simple SIR model \eqref{eq:basic_SIR_ODE_matrix} can be extended with additional compartments or subdividing existing ones in a particular way. One such model is the S3I3R model, which partitions \textit{infected} into the three sub-compartments, \textit{infected} ($I_1$), \textit{hospitalized} ($I_2$) and intensive care unit \textit{ICU} ($I_3$). Furthermore, \textit{removed} is split into \textit{recovered} ($R_1$), \textit{vaccinated} ($R_2$) and \textit{dead} ($R_3$). The dynamical flow between compartments of the S3I3R model is illustrated in Figure \ref{fig:S3I3R_flowchart}. The flow is governed by the system of seven coupled ODEs presented in \eqref{eq: expanded_SIR_ODE}, which is identified by eight parameters, each describing the flow from compartment $A$ to $B$, defined as the rate at which the population moves from $A$ to $B$, except for $\beta$, which is a combined measure of both interactivity in the population and the infectivity of the disease.

\begin{align} \label{eq: expanded_SIR_ODE}
    \frac{\diff S}{\diff t} &= -\left(\frac{\beta I_1}{N}+\frac{\tau}{S+I_1+R_1}\right) S, \nonumber \\
    \frac{\diff I_1}{\diff t} &= \frac{\beta I_1}{N} S - \left(\gamma_1 + \frac{\tau}{S + I_1 + R_1} + \phi_1\right) I_1, \nonumber \\
    \frac{\diff I_2}{\diff t} &= \phi_1 I_1 - (\gamma_2 + \phi_2)I_2, \nonumber \\
    \frac{\diff I_3}{\diff t} &= \phi_2 I_2 - (\gamma_3 + \theta) I_3, & t\geq0.\\
    \frac{\diff R_1}{\diff t} &= \gamma_1 I_1 + \gamma_2 I_2 + \gamma_3 I_3 - \frac{\tau }{S+I_1+R_1}R_1, \nonumber \\
    \frac{\diff R_2}{\diff t} &= \tau, \nonumber \\
    \frac{\diff R_3}{\diff t} &= \theta I_3, \nonumber 
\end{align}

Similarly to the basic SIR model, the system \eqref{eq: expanded_SIR_ODE} is parameter-linear and can be written in matrix form as 
\begin{equation}\label{eq:S3I3R_matrix}
    \frac{\diff {\bf x}}{\diff t} = A{\mat\omega}, \quad t \geq 0,
\end{equation}
\begin{equation*}
    {\mat x} = 
    \begin{bmatrix}
        S \\ I_1 \\ I_2 \\ I_3 \\ R_1 \\ R_2 \\ R_3
    \end{bmatrix}, \quad
    \mat{A} =
    \begin{bmatrix}
        \frac{-SI_1}{N} & 0 & 0 & 0 & \frac{-S}{S+I_1+R_1} & 0 & 0 & 0 \\
        \frac{SI_1}{N}  & -I_1 & 0 & 0 & \frac{-I_1}{S+I_1+R_1} & 0 & -I_1 & 0 \\
        0 & 0 & -I_2 & 0 & 0 & 0 & I_1 & -I_2 \\
        0 & 0 & 0 & -I_3 & 0 & -I_3 & 0 & I_2 \\
        0 & I_1 & I_2 & I_3 & \frac{-R_1}{S+I_1+R_1} & 0 & 0 & 0 \\
        0 & 0 & 0 & 0 & 1 & 0 & 0 & 0 \\
        0 & 0 & 0 & 0 & 0 & I_3 & 0 & 0
    \end{bmatrix}, \quad
    {\mat\omega} = \begin{bmatrix}
        \beta \\ \gamma_1 \\ \gamma_2 \\ \gamma_3 \\ \tau \\ \theta \\ \phi_1 \\ \phi_2
    \end{bmatrix}.
\end{equation*}

Due to the assumption of constant population, each column of $\mat{A}$ sums up to 0, implying that any flow out of one compartment must enter another. Notable is that each compartment only has 1-4 connections to other compartments, which in turn makes the system matrix $\mat{A}$ sparse.

\subsubsection{Estimating Time-constant Parameters}\label{sec: PIR_est_const_params}

A fractional population is initialized such that 1 in 10,000 is infected, and the remaining population is susceptible to the disease using the initial condition
\begin{align}
    \mat{x}_0 
    &= \begin{bmatrix} S_0 & I_0 & R_0 \end{bmatrix}^T 
    = \begin{bmatrix} 0.9999 & 0.0001 & 0 \end{bmatrix}^T.
    \label{eq: SIR const initial}
\end{align}
We made the following arbitrary choice for the transmission coefficient and recovery rate, 
\begin{align}
    {\mat\omega} 
    &= \begin{bmatrix} \beta & \gamma \end{bmatrix}^T 
    = \begin{bmatrix} \frac{1}{2} & \frac{1}{3} \end{bmatrix}^T \;\; \frac{1}{\text{day}}.
    \label{eq: SIR const params}
\end{align}

\begin{figure}[!h]
  \centering
  \subfigure[SIR simulation]{%
    \includegraphics{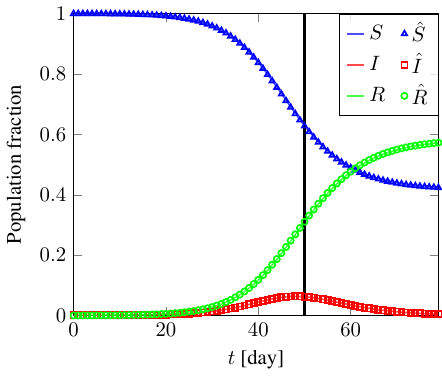}
    \label{fig: PIR_SIR_synt_const_states}
  }
  \hspace{.5cm} 
  \subfigure[S3I3R simulation]{%
    \includegraphics{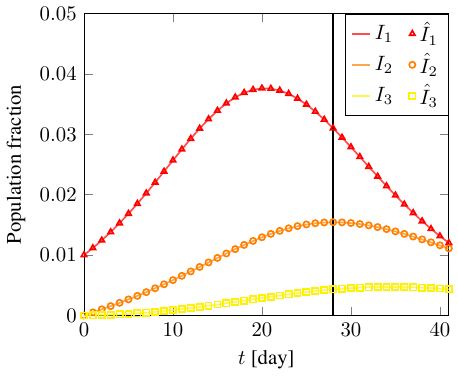}
    \label{fig: PIR_S3I3R_synt_const_states}
  }
  \caption{SIR and S3I3R simulations using PIR estimated parameters. Vertical lines indicate the train/test split. Curves and markers are simulated using the true and estimated parameters respectively.}
  \label{fig: PIR_synt_const_states}
\end{figure}

Substituting the parameters \eqref{eq: SIR const params} into the ODE \eqref{eq:SIR_ODE}, make for an initial value problem, with initial conditions \eqref{eq: SIR const initial}, which is solved numerically using the ERK4 method\cite{Butcher2016NumericalEquations}. The solutions ${\mat x}_i$ for the first days $i=1,...,80$ are stored as synthetic simulated daily data. Using the first 50 daily states, i.e. ${\mat x}_i$ for $i \in I=\{1,2,\dots,50\}$, the optimal set of parameters ${\mat\omega}_I^*$ is estimated over this temporal domain by using the OLS formula \eqref{eq:OLS_multi}. For explicity, the system matrix ${\mat A}_i$ and derivative $\diff {\mat x}_i/ \diff t$ are computed according to \eqref{eq:basic_SIR_ODE_matrix} and \eqref{eq:central_difference}, given ${\mat x}_{i-1},{\mat x}_i,{\mat x}_{i+1}$ for all $i \in I$. The relative errors between the estimated optimal parameters $\omega_{I}^*$, and the parameters used to simulate the epidemic, are stated in Table \ref{tab: rel_error_PIR}. To visualize the accuracy of the method, the pandemic is simulated again, this time using the reconstructed parameters ${\mat \omega}_I^*$ instead. Both the simulated data and the synthetic data are illustrated in Figure \ref{fig: PIR_synt_const_states}.

The same experiment is also conducted for the S3I3R model \eqref{eq: expanded_SIR_ODE}. The initial conditions are defined as,
\begin{align}
    {\mat x}_0 &= \begin{bmatrix}
        S_0 & I_{1,0} & I_{2,0} & I_{3,0} & R_{1,0} & R_{2,0} & R_{3,0}
    \end{bmatrix}^T \nonumber \\
    &= \begin{bmatrix}
        0.9999 & 0.0001 & 0.0 & 0.0 & 0.0 & 0.0 & 0.0
    \end{bmatrix}^T. \label{eq: S3I3R const initial}
\end{align}
with model parameters,
\begin{align}
    {\mat\omega} 
    &= \begin{bmatrix}
        \beta & \gamma_1 & \gamma_2 & \gamma_3 & \phi_1 & \phi_2 & \theta & \tau
    \end{bmatrix}^T \nonumber \\
    &= \begin{bmatrix}
        \frac{1}{2} & \frac{1}{3} & \frac{1}{20} & \frac{1}{20} & \frac{1}{20} & \frac{1}{20} & \frac{1}{10} & 0
    \end{bmatrix}^T \frac{1}{\text{day}},
    \label{eq: S3I3R const params}
\end{align}
and the initial value problem is solved with ERK4, and the solutions ${\mat x}_i$ are stored for the first $i=1,...42$ days. Using the first 28 of the 42 states, i.e. ${\mat x}_i$ for $i \in I=\{1,2,\dots,28\}$, the estimated parameters ${\mat\omega}_I^*$ are computed for this time period, using the least squares formula \eqref{eq:OLS_multi}. Once again, the system matrix ${\mat A}_i$ and the derivative $d{\bf x}_i/dt$ are computed according to \eqref{eq:S3I3R_matrix} and \eqref{eq:central_difference}, given ${\bf x}_{i-1},{\bf x}_i,{\bf x}_{i+1}$ for all $i \in I$. The relative errors between the estimated parameters ${\mat\omega}_{I}^*$, and the parameters used to simulate the epidemic, are presented in Table \ref{tab: rel_error_PIR}. The epidemic is simulated once again using the reconstructed parameters ${\mat\omega}_I^*$. The simulation and synthetic data are shown in Figure \ref{fig: PIR_synt_const_states}.

\begin{table}[t!]
    \centering
    \begin{tabular}{c|c}
        \toprule
        \multicolumn{2}{c}{\bf SIR} \\
        \midrule
        Estimator & Relative Error \\
        \midrule
        ${\hat \beta}$   & $1.84\cdot10^{-4}$ \\
        ${\hat \gamma}$ & $6.48\cdot10^{-4}$ \\
        \bottomrule
    \end{tabular}
    \hspace{1cm}
    \begin{tabular}{c|c}
        \toprule
        \multicolumn{2}{c}{\bf S3I3R} \\
        \midrule
        Estimator & Relative Error \\
        \midrule
        ${\hat \beta}$ & $3.78\cdot10^{-4}$ \\
        ${\hat \phi_1}$  & $1.11\cdot10^{-3}$ \\
        ${\hat \phi_2}$  & $2.36\cdot10^{-4}$ \\
        ${\hat \theta}$  & $2.22\cdot10^{-4}$ \\
        \bottomrule
    \end{tabular}
    \caption{Relative errors for the PIR estimated parameters.}
    \label{tab: rel_error_PIR}
\end{table}


A single successful example of parameter estimation is insufficient to validate this as a general method. Thus, the SIR model experiment is repeated by sampling many pairs of values of $\beta$ and $\gamma$.

We proceed from fractional populations to actual population sizes to see whether this would cause numerical instabilities for some choices of $\beta$ and $\gamma$. The population is initialized as ${\mat x}_0 = [5600000, 100000, 0]^T$ people, and then all combinations of $\beta \in [0, 0.5]\text{day}^{-1}$ and $\gamma \in [0, 0.3]\text{day}^{-1}$.

\begin{figure}[H]
    \centering
    \includegraphics{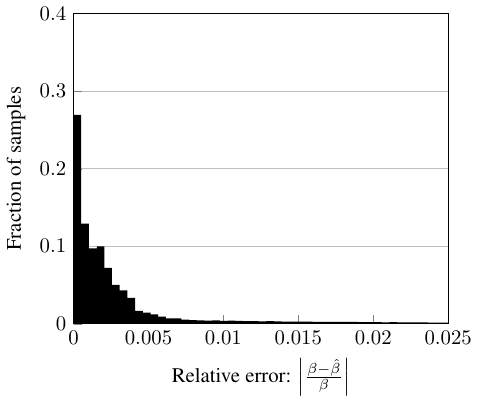}
    \includegraphics{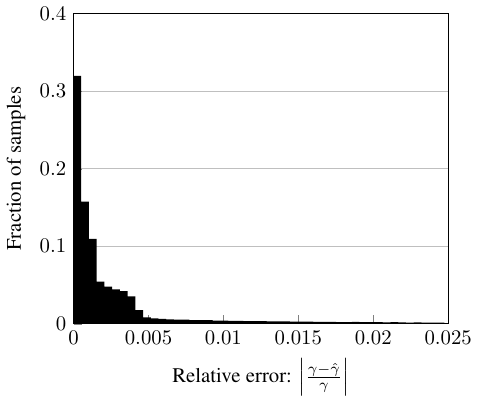}
    \caption{Absolute relative error distribution of $\beta$ (Left) and $\gamma$ (Right) estimates for systems sampled random uniform from domain $(\beta,\gamma)\in [0, 0.5]\frac{1}{\text{day}} \times [0, 0.3]\frac{1}{\text{day}}$.}
    \label{fig:basic_errors}
\end{figure}

The histograms in Figure \ref{fig:basic_errors} show the distribution of the relative error of each parameter - for both parameters the relative error is always below 5\%, but most are below 0.5\%. 

Next, we increase the complexity of the model significantly, using the S3I3R illustrated in figure \ref{fig:S3I3R_flowchart}.

This model has eight parameters of which only seven are estimated since the vaccination parameter ($\tau$) generally does not have to be estimated, and also has a completely different order of magnitude. Due to the amount of additional parameters present within this model, exhaustively evaluating all potential combinations is computationally infeasible. Consequently, a strategy of sampling random combinations of parameters $\omega \in \boldsymbol{D} \subset \mathbb{R}^7$ within a predefined domain is employed. The domain is based on prior knowledge on these parameters~\cite{Mller2020TekniskModellerne} and is set as,
\begin{equation*}
    \boldsymbol{D} = [0, 0.5]\frac{1}{\text{day}} \times [0, 0.3]\frac{1}{\text{day}} \times [0, 0.3]\frac{1}{\text{day}} \times [0, 0.3]\frac{1}{\text{day}} \times [0, 0.03]\frac{1}{\text{day}} \times [0, 0.3]\frac{1}{\text{day}} \times [0, 0.5]\frac{1}{\text{day}}.
\end{equation*}

Using initial values ${\mat x}_0 = [5600000, 100000, 1000, 10, 0, 0, 0]$ individuals, and simulating 100 time steps, we sample 10000 times from $\boldsymbol{D}$. The results are presented Table \ref{tab:errors}.

\begin{table}[H]
    \centering
    \begin{tabular}{l|ccccc}
        \toprule
        \textbf{Relative Error} & $< 1.00$ & $< 0.50$ & $< 0.10$ & $< 0.05$ & $< 0.01$ \\
        \midrule
        Max & 0.86 & 0.80 & 0.50 & 0.36 & 0.10 \\
        Mean & 0.97 & 0.95 & 0.88 & 0.85 & 0.74 \\
        Max (normalizing) & 0.97 & 0.93 & 0.64 & 0.54 & 0.27 \\
        Mean (normalizing) & 0.99 & 0.99 & 0.93 & 0.81 & 0.56 \\
        \bottomrule
    \end{tabular}
    \caption{Fraction of the 10000 simulated samples below different error bounds. As an example, the mean relative error is less than 0.10 in 88\% of simulated samples.}
    \label{tab:errors}
\end{table}

It is seen from Table \ref{tab:errors} that the maximum errors are generally smaller when using normalization, whereas the mean error tend to be smaller when not using normalization. Worth mentioning is that in around 90\% of the samples, the mean error was smaller than 10\% regardless of normalization.

\subsubsection{Estimating Time-varying Parameters}\label{PIR_est_var_params}
Generally, the infection rate $\beta$ is assumed to be constant in most SIR-type models~\cite{Cooper2020ACommunities, Chen2020APersons}. A problem arises when fitting models with a constant infection rate to real data, where the actual dynamics of transmission/infection change wildly over a period. In the SIR-type models, both the viral properties (contagiousness, mediums of transmission) and the social component (level of physical human interaction) are captured in the single parameter $\beta$. Denmark and many other countries have adopted the approach of (micro)managing the COVID-19 pandemic with different lockdown policies and mandates, aiming to regulate physical human interactions in the country. Such policies will, hopefully, result in a reduction of infection rates, which is reflected in the parameter $\beta$.

\begin{figure}[H]
  \centering
  \subfigure[SIR model estimation accuracy]{%
    \includegraphics{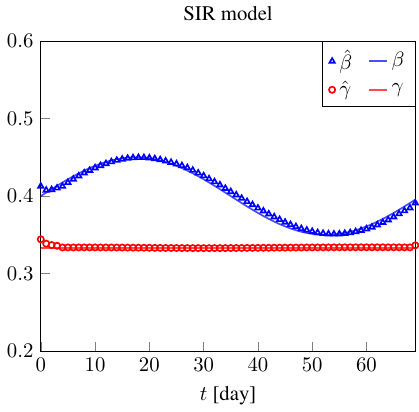}
    \label{fig: PIR_SIR_synt_vary_params}
  }
  \hspace{1cm} 
  \subfigure[S3I3R model estimation accuracy]{%
    \includegraphics{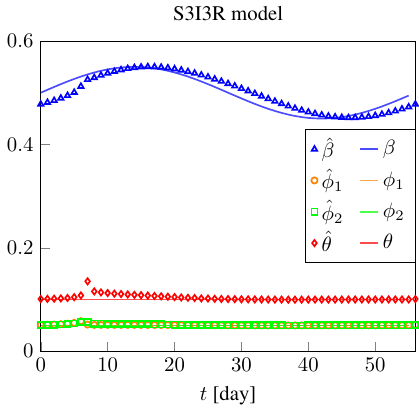}
    \label{fig: PIR_S3I3R_synt_vary_params}
  }
  \caption{SIR and S3I3R estimation accuracy using varying parameters.}
  \label{fig: PIR_synt_vary_params}
\end{figure}

In this experiment, the synthetic data is generated using a varying infection rate. That is, parameters and initial conditions in both models are defined as in Section \ref{sec: PIR_est_const_params}, except for $\beta$ which will follow some arbitrary curve through time. For this example, the infection rate is implemented as a sinusoidal wave $\beta(t)$, with a period of $T$ days and an amplitude of 0.05 centered around the value $\overline{\beta}$. 
\begin{equation}
    \beta(t) = 0.05\sin\left(\frac{2\pi t}{T}\right)+\overline{\beta}.
\end{equation}
To simulate an epidemic using $\beta(t)$, one of the two models, \eqref{eq:SIR_ODE} or \eqref{eq: expanded_SIR_ODE}, is solved with an RK4-solver~\cite{Butcher2016NumericalEquations}, where parameters are updated in each iteration $i$, such that $\beta\leftarrow \beta(t_i)$. This generates the synthetic data $x_i$ at time $t_i$ for all $i\in I=\{1,...,n\}$. The parameters $\hat{\omega}_i$ are then estimated from the generated data for all indices $I$ with the PIR equation \eqref{eq: OLS}, by using the corresponding model \eqref{eq:basic_SIR_ODE_matrix} or \eqref{eq:S3I3R_matrix}.
To evaluate how well the parameters are estimated, the \textit{Mean Relative Error} (MRE) is used. It is defined as follows: 
\begin{equation}
    \text{MRE}_p = \frac{1}{n}\sum_i^n\frac{p(t_i)- \hat{p}(t_i)}{p(t_i)} \quad (\textit{Mean Relative Error}),
\end{equation}
for an arbitrary parameter $p$, and note if a parameter is constant in the simulation, then $p(t_i)=p$, which will be the case for all parameters except $\beta$. The experiment is conducted for the basic model in Section \ref{sec: SIR}, with $n=T=70 \text{day}$, and $\overline{\beta}=0.4\text{day}^{-1}$ and for the expanded model in Section \ref{sec: S3I3R} with $n=T=56\text{day}$, and $\overline{\beta}=0.4 \text{day}^{-2}$. In both cases, the data is evenly spread such that $t \in \{1,\dots,T\}$. The reconstructed parameters $\hat{\omega}_i$ for $i \in I$ are visualized in Figure \ref{fig: PIR_synt_vary_params}, along with the parameters ${\mat\omega}_i$ used to create the simulation. The mean relative errors of the estimated parameters are shown in Table \ref{tab: RE_PIR}. 

\begin{figure}[H]
  \centering
  \subfigure[SIR model with noise]{%
    \includegraphics{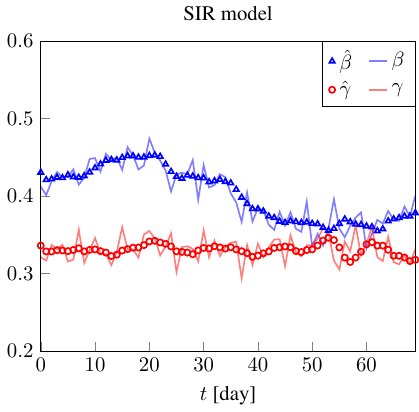}
    \label{fig:PIR_SIR_synt_vary_params_noise}
  }
  \hspace{1cm} 
  \subfigure[S3I3R model with noise]{%
    \includegraphics{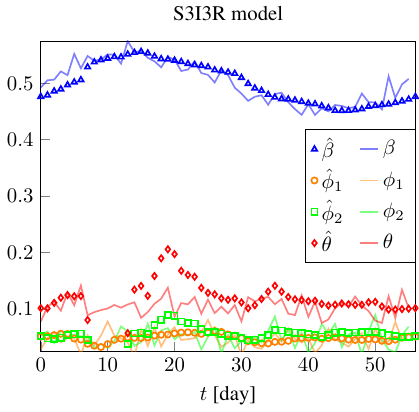}
    \label{fig:PIR_S3I3R_synt_vary_params_noise}
  }
  \caption{Parameter estimation for SIR and S3I3R models with varying parameters and noisy data.}
  \label{fig:PIR_synt_vary_params_noise}
\end{figure}

\begin{table}[H]
    \centering
    \begin{tabular}{c|c}
        \toprule
        \multicolumn{2}{c}{\bf SIR} \\
        \midrule
        Estimator & Relative Error \\
        \midrule
        ${\hat \beta}$   & $1.05\cdot10^{-2}$ \\
        ${\hat \gamma}$  & $1.00\cdot10^{-2}$ \\
        \bottomrule
    \end{tabular}
    \hspace{1cm}
    \begin{tabular}{c|c}
        \toprule
        \multicolumn{2}{c}{\bf S3I3R} \\
        \midrule
        Estimator & Relative Error \\
        \midrule
        ${\hat \beta}$    & $2.99\cdot10^{-2}$ \\
        ${\hat \phi_1}$   & $2.97\cdot10^{-2}$ \\
        ${\hat \phi_2}$   & $3.82\cdot10^{-2}$ \\
        ${\hat \theta}$   & $6.96\cdot10^{-2}$ \\
        \bottomrule
    \end{tabular}
    \caption{Relative error for estimation of time-varying parameters using PIR.}
    \label{tab: RE_PIR}
\end{table}

It appears that the PIR method can estimate the temporal changes in the model parameters over time. Another experiment is therefore conducted with the addition of some noise on the parameters. The noise is drawn randomly from a zero-centered normal distribution with standard deviation 0.01 and is added to each of the parameters before generating the synthetic data. The reconstructed parameters are shown in Figure \ref{fig:PIR_synt_vary_params_noise} along the parameters used to generate the data.


\subsubsection{Real Data}\label{sec:SIRrealdata}

It is now attempted to estimate the parameters from data collected from the COVID-19 pandemic in Denmark. It is assumed that the infection rate ($\beta$) varies over time, and thus this experiment is almost equivalent to that of Section \ref{PIR_est_var_params}, using the basic SIR model. 

\begin{figure}[H]
    \centering
    \includegraphics{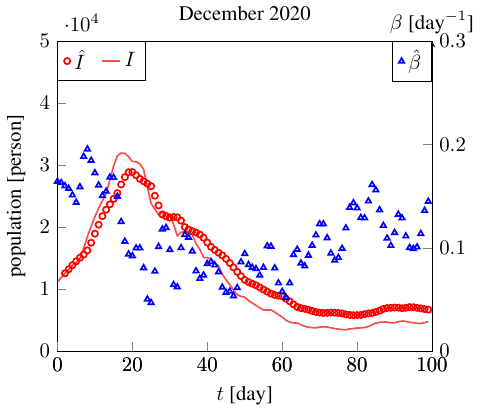}
    \includegraphics{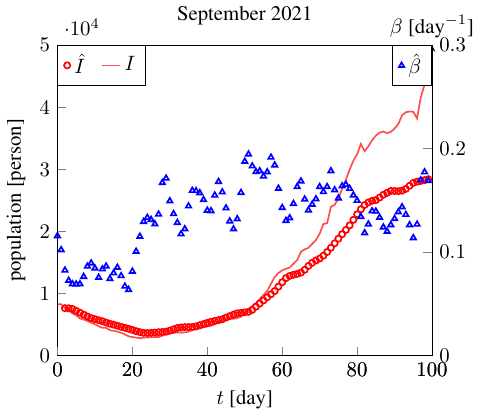}
    \caption{SIR model fitted to real data. The parameter $\beta$ is estimated using the PIR method for each observation. Then the model propagates forward through time using the initial $S_0,I_0,R_0$ values, and the estimated parameters $\beta$. $\beta$ is also illustrated on a different axis.}
    \label{fig: realdata}
\end{figure}

The only difference is the data used, which is collected from WHO (see Appendix \ref{sec: note on datatrans}), and the experiments will be conducted for 100 days of data. First, starting from December 1st 2020, and then starting from September 1st 2021. The parameter for day $i$ is computed using data from the preceding 14 days $\{i-13, i-12,\dots,i-1, i\}$. The computed varying parameters $\hat{\mat\omega}_i$ for $i \in I$, will be used to simulate the pandemic using the ERK4~\cite{Butcher2016NumericalEquations} solver for the temporal integration, starting from the above-mentioned dates. The infected component from the simulation is illustrated in Figure \ref{fig: realdata}, along with the actual infected data provided by WHO. The development of $\hat{\beta}$ is shown in the same figure.

As mentioned, it is often necessary to incorporate the preceding time steps to ensure that the matrix $A$ is tall. For real-world data, where the infection rate $\beta$ may vary over time, this temporal window represents a relatively short period, in which $\beta$, or any model parameter, is assumed to remain constant. Selecting an interval too wide may introduce a smoothing effect, potentially obscuring meaningful dynamics. Contrary, a too narrow interval can lead to numerical instabilities and significant fluctuations in successive parameter estimates.

\section{Experiments with PINNs on epidemiological models}
\label{sec:PINN}
The purpose of testing the PINN method for parameter estimation is to provide an alternative to the OLS method. The experiments are designed to be comparable to those performed with OLS, such that a fair analysis can be conducted.  

\subsection{Estimating Time-constant Parameters}\label{estimating constant params pinn}
Similar to section \ref{sec: PIR_est_const_params}, this experiment considers the estimation of model parameters from synthetic data generated by constant parameters. The data is equivalent to that which was generated and used in Section \ref{sec: PIR_est_const_params}. The only difference of course, being the method of estimating parameters ${\mat\omega}_I^*$.

\begin{table}[H]
    \centering
    \begin{tabular}{c|c}
        \toprule
        \multicolumn{2}{c}{\bf SIR} \\
        \midrule
        Estimator & Relative Error \\
        \midrule
        ${\hat \beta}$   & $4.68\cdot10^{-3}$ \\
        ${\hat \gamma}$ & $4.44\cdot10^{-2}$ \\
        \bottomrule
    \end{tabular}
    \hspace{1cm}
    \begin{tabular}{c|c}
        \toprule
        \multicolumn{2}{c}{\bf S3I3R} \\
        \midrule
        Estimator & Relative Error \\
        \midrule
        ${\hat \beta}$ & $2.48\cdot10^{-3}$ \\
        ${\hat \phi_1}$  & $1.86\cdot10^{-2}$ \\
        ${\hat \phi_2}$  & $3.24\cdot10^{-2}$ \\
        ${\hat \theta}$  & $9.51\cdot10^{-2}$ \\
        \bottomrule
    \end{tabular}
    \caption{Relative error of PINN parameter estimates.}
    \label{tab: rel_error_PINN}
\end{table}

\begin{figure}[H]
  \centering
  \subfigure[SIR model prediction]{%
    \includegraphics{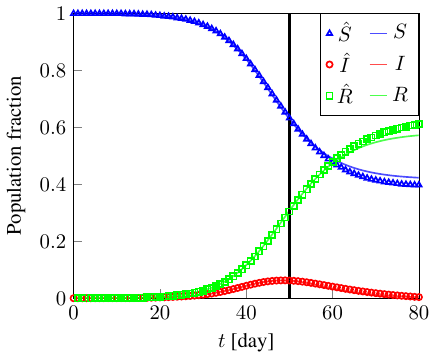}
    \label{fig:sir_const}
  }
  \hspace{1cm} 
  \subfigure[Parameter convergence: SIR model]{%
    \includegraphics{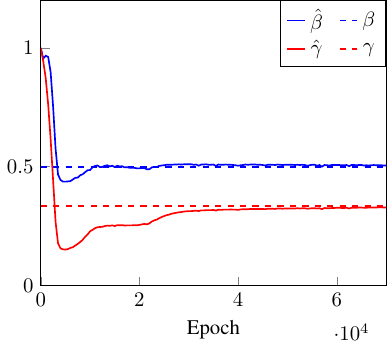}
    \label{fig:sir_params}
  }
  \\ 
  \subfigure[S3I3R model prediction]{%
    \includegraphics{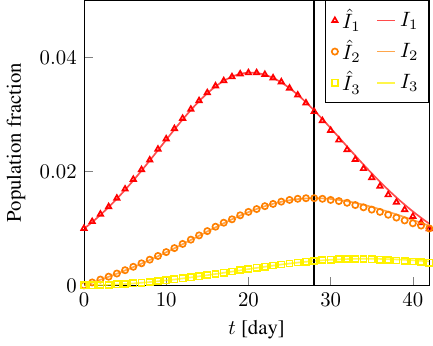}
    \label{fig:s3i3r_const}
  }
  \hspace{1cm} 
  \subfigure[Parameter convergence: S3I3R model]{%
    \includegraphics{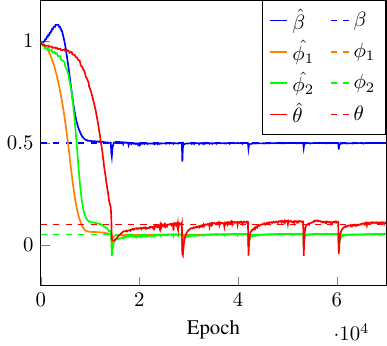}
    \label{fig:s3i3r_params}
  }
  \caption{PINN results for the parameter estimation of the model parameters in the models SIR and S3I3R.}
  \label{fig:SIR_and_S3I3R_PINN}
\end{figure}

The parameter convergence plots in Figures \ref{fig:sir_params} and \ref{fig:s3i3r_params} are presented in terms of epochs, where each epoch represents a complete pass of the PINN through the training dataset. Various initial conditions and training time intervals were considered and tested. The plots included in this paper represent the model configurations that, during testing, yielded the most favorable convergence. Various configurations were explored, but many resulted in significantly worse performance. Further details on these experiments are provided in the Appendix \ref{section: note on IC}. 

\subsection{Estimating Time-varying Parameters}\label{estimating varying params pinn}
The experiment in Section \ref{PIR_est_var_params} is now repeated using a PINN to estimate the $\beta(t)$ values. For each day $t_i$, a particular slice of the data is used as training data:

Let $d$ be the number of extra days used as training data for a particular day's estimate. This is analogous to the number of equations used in the over-determined system when using the PIR method. For each day $t_i$, the training data then consists of data from the days in the interval $[t_i-d,t_i]$. That is, for each day, the previous $d$ days are considered as training data (current day included).

\begin{table}[H]
    \centering
    \begin{tabular}{c|c}
        \toprule
        \multicolumn{2}{c}{\bf SIR} \\
        \midrule
        Estimator & Relative Error \\
        \midrule
        ${\hat{\beta}}$   & $1.28\cdot10^{-1}$ \\
        ${\hat{\gamma}}$  & $2.81\cdot10^{-2}$ \\
        \bottomrule
    \end{tabular}
    \hspace{1cm}
    \begin{tabular}{c|c}
        \toprule
        \multicolumn{2}{c}{\bf S3I3R} \\
        \midrule
        Estimator & Relative Error \\
        \midrule
        ${\hat{\beta}}$   & $2.79\cdot10^{-1}$ \\
        ${\hat{\phi}_1}$  & $1.53\cdot10^{0}$ \\
        ${\hat{\phi}_2}$  & $1.52\cdot10^{0}$ \\
        ${\hat{\theta}}$  & $8.27\cdot10^{-1}$ \\
        \bottomrule
    \end{tabular}
    \caption{Relative Error for varying parameters using PINNs.}
    \label{tab: PINNRE}
\end{table}

\begin{figure}[H]
  \centering
  \subfigure[SIR model]{%
    \includegraphics{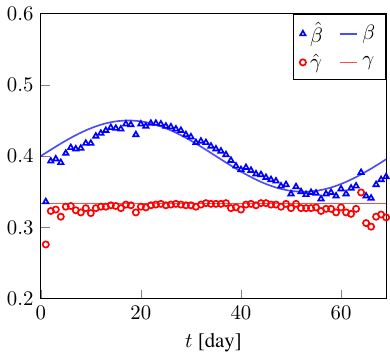}
    \label{fig: SIR_varying_params}
  }
  \hspace{1cm} 
  \subfigure[S3I3R model]{%
    \includegraphics{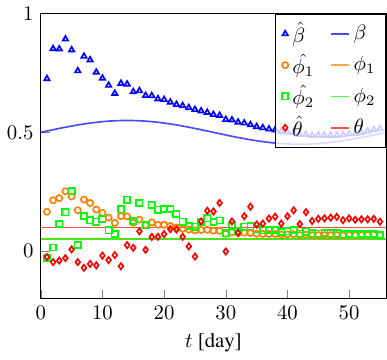}
    \label{fig: S3I3R_varying_params}
  }
  \caption{SIR and S3I3R parameter estimation accuracy using varying parameters.}
  \label{fig: est_accuracy}
\end{figure}


The Relative Error is calculated for each of the parameters estimated in each model.



\section{Experiment with PIR for partial differential equations}\label{sec:PIRexpPDE}

\subsection{Estimating the Reynolds Number from Vortex Shedding Data}

In this section, we consider a classical case of confined fluid flow past a cylinder (cf. review \cite{Nguyen2023ACylinders}) and use the public data included in the book \cite{Brunton2019Data-DrivenControl}. Such data were used for non-linear model identification \cite{Brunton2016DiscoveringSystems} and have been reported to be produced using the immersed boundary method \cite{Taira2007TheApproach,Colonius2008AConditions}. The fluid flow is described by the dimensionless incompressible Navier-Stokes equations stated as
\begin{subequations}
\begin{align}
\frac{\partial \mathbf{u}}{\partial t} + \mathbf{u} \cdot \nabla \mathbf{u} &= - \nabla p + \frac{1}{Re} \nabla^2 \mathbf{u}, \label{eq:momentumeqNSE} \\
\nabla \cdot \mathbf{u} &= 0,
\end{align}
\label{eq:INSE}
\end{subequations}
where $\mathbf{u} = (u,v)$ is the velocity field and $p$ is the pressure. The Reynolds number is defined as
\begin{align}
\text{Re} = \frac{U L}{\nu},
\end{align}
where $U$ is the characteristic velocity, $L$ is the characteristic length scale, and $\nu=\mu/\rho$ is the kinematic viscosity, $\mu$ is the dynamic viscosity and $\rho$ is the density of the fluid. The Reynolds number characterizes the flow regime, with low values indicating laminar flow and high values indicating turbulent flow. The fluid flow is setup using the following dimensionless quantities:
$U=1$, $V=0$, $Re=100$.

Furthermore, with only velocity and vorticity data available, our aim is to learn from the data what the correct Reynolds number is for the fluid flow. Since fluid flow can be described by the incompressible Navier-Stokes equations \eqref{eq:INSE} we recognize that the equations are parameter-linear wrt. the Reynolds number. The idea is to use the PIR method to estimate the Reynolds number, however, the INS are expressed in terms of pressure which is unknown. Instead, we derive the evolution equation for vorticity, starting from the definition of vorticity
\begin{align}
\omega \equiv \nabla \times {\bf u}.
\label{eq:vorticityeq}
\end{align}

Hence, by taking the curl operator of \eqref{eq:momentumeqNSE} we find an equation that governs the temporal evolution of the vorticity 
\begin{align}
\omega_t + \mathbf{u} \cdot \nabla \omega = \frac{1}{Re} \nabla^2 \omega.
\label{eq:vorticity}
\end{align}
Remark, the pressure terms cancel since mixed partial derivatives commute for a smooth scalar field (cf. Schwarz theorem) such as pressure, and hence \eqref{eq:vorticity} is expressed in a form that leads to a causal relationship governed by a physical model that can be used with the available vorticity and velocity field data.

Next, a bit of thought needs to be put into processing the snapshot data that are available. We need to determine the time step size and the spatial increments in the grid used for storing the data.

The Strouhal number $St$ is a dimensionless number that relates the vortex shedding frequency to the flow speed and characteristic length
\begin{align*} 
St = \frac{f D}{U},
\end{align*}
where $f$ is the vortex shedding frequency, $D$ is the diameter of the cylinder, $U$ is the free-stream velocity. At $Re = 100$, the Strouhal number is approximately $0.164$. To solve for the vortex shedding period $T$, we use $t= 1 / T$ (since U = D = 1). We find $T = 1 / St = 1 / 0.164 = 6.10$ time units. This is the duration of one shedding period. The data contains snapshots over a total of $N_p=5$ shedding periods. From this we can deduce the time-step size assuming that we have evenly spaced snapshots across this period, then the time step is determined as 
$
\Delta t = N_p T / N = N_p / (St \cdot N_s)
$
 where $N_s=151$ is the number of full-state snapshots in the data sets. Hence, a temporal step size between the snapshots 
$
\Delta t = \frac{N_p\cdot T}{N_s} = \frac{5\cdot 6.10}{151} \approx 0.202$ time units is estimated. The spatial grid arrangement is such that there are $(N_x,N_y)=(449,199)$ nodes in a spatial grid, the cylinder diameter covering 25 grid nodes. Hence, the grid increments are $\Delta x=\Delta y=0.02$.

\begin{figure}[H]
    \centering
    \subfigure[Vorticity field snapshot at sample 70.]{%
        \includegraphics[width=0.70\textwidth]{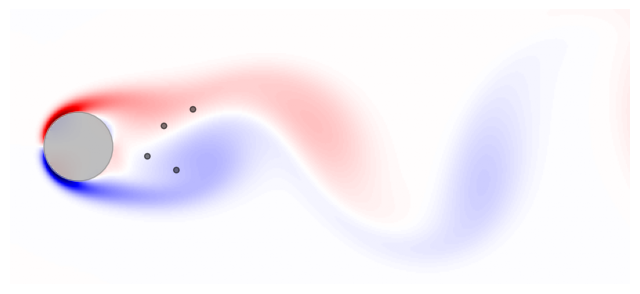}   
    }
    \subfigure[Convergence of Reynolds number.]{%
        \includegraphics[width=0.45\textwidth]{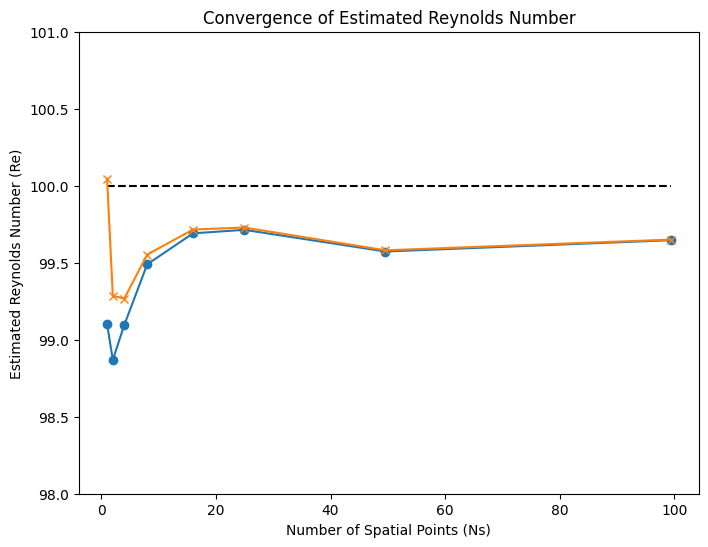}
    }
    \subfigure[Relative Error of Reynolds number.]{%
        \includegraphics[width=0.45\textwidth]{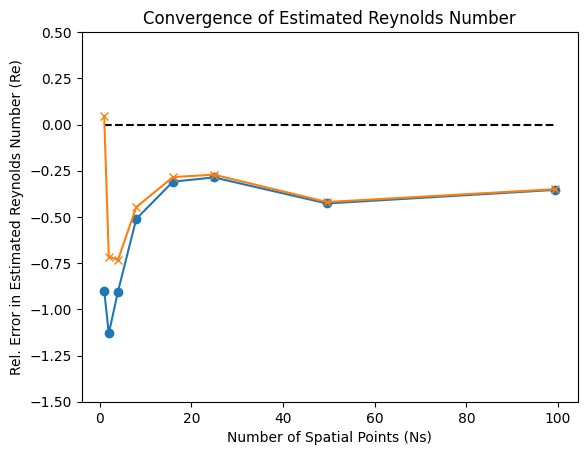}
    }
    \caption{Flow past a cylinder. a) Snapshot of the vorticity field with 4 random sensor positions (black dots) in the wake region illustrated. b) Convergence in estimated Reynolds number as a function of spatial points (sensors) using PIR without ridge regression (dots) and with ridge regression (crosses). c) Relative errors [\%] as a function of spatial points (sensors) using PIR without ridge regression (dots) and with ridge regression (crosses). }
    \label{fig:PDEcylinderflow}
\end{figure}

To use the PIR method, we need to estimate from the snapshots the temporal and spatial derivatives in \eqref{eq:vorticityeq} for a set of $N_s$ sensor points in space. For each of these sensor points we can extract the time series data across the snapshots available, and from these estimates, using second-order accurate central finite difference, the temporal and spatial gradients of first and second order, e.g. 
\begin{align*}
\frac{\partial \omega}{\partial t} \approx \frac{\omega^{n+1} - \omega^n}{t_{n+1}-t_n}, \quad 
\frac{\partial \omega}{\partial x} \approx \frac{\omega^{n}_{i+1,j} - \omega^n_{i-1,j}}{x_{i+1,j}-x_{i-1,j}}, \quad 
\frac{\partial^2 \omega}{\partial x^2} \approx \frac{\omega^{n}_{i+1,j} + 2 \omega^{n}_{i,j} - \omega^n_{i-1,j}}{(x_{i+1,j}-x_{i-1,j})^2}, 
\end{align*}
and similarly in the $y$-direction, where $\omega(t_n)\equiv \omega^n$, $x_{i,j}=i\Delta x+x_0$, $y_{i,j}=j\Delta y+y_0$, $t_n = n\Delta t$, $n=1,...,N$. 
By collecting the total of $s=1,...,N_s$ time series, this leads to a linear system
\begin{align}
\left[  
\begin{array}{c} 
| \\
\left[  
\begin{array}{c} 
\tfrac{\partial \omega}{\partial t} + {\bf u}\cdot \nabla \omega
\end{array} 
\right]_s
\\
|  
\end{array} 
\right] = 
\left[  
\begin{array}{c} 
|  \\
 \left[  
\begin{array}{c} 
[\Delta \omega]_s
\end{array} 
\right] 
\\
|
\end{array} 
\right] 
\left[  
\begin{array}{c} 
\frac{1}{Re}
\end{array} 
\right]  \quad \leftrightarrow \quad {\bf b} = {\bf A} {\bf x}.
\end{align} 
To evaluate convergence and performance of the selected sensor data, we vary the number of sensor positions randomly restricted to the wake region of the cylinder, since we expect a favorable signal-to-noise ratio in this area. For each of the $N_s$ sensors, we solve the PIR problem 20 times using different random seeds and take the average of the estimated Reynolds numbers and store these, cf. Figure \ref{fig:PDEcylinderflow} a). We provide a snapshot (data sample 70) of the random sensor positions for a setup with 4 sensors (black dots) in the wake region of the cylinder (gray shaded region) in Figure \ref{fig:PDEcylinderflow} a). The convergence results are presented in Figures \ref{fig:PDEcylinderflow} b) and c), and show that errors are below 1.5 percent on average, and with only 4 sensors in the wake region we achieve less than 1 percent error using the PIR method with a trend towards just below 0.5 percent error in the estimated Reynolds numbers. When using ridge regression (cf. Appendix \ref{sec:regLSQproof}) with a non-optimized setting of  $\lambda=1000$ it is shown that it is possible to improve the accuracy of the parameter estimation when few sensors are selected. 

%


\section{Conclusion}
A physics-informed regression (PIR) method that relies on least-square regression as a solution method is introduced to estimate parameters of time-dependent parameter-linear dynamic models. PIR comes with the advantage of relying on a closed-form expression based on the least square method. The application of PIR for parameter estimation include different nonlinear dynamical systems of ODE and PDE types. Numerical experiments highlight how to use PIR for ODE systems such as the Lorenz system and epidemic modeling, where for the latter it was found that the PIR method outperforms the PINN method for parameter estimation of ODEs. Although both methods succeeded in estimating the time-constant parameters to an acceptable degree, PIR was found to perform better at estimating the time-varying parameters. Crucially, PIR outperformed PINN in the time taken to calculate each estimate highlighting the efficiency of PIR. The ability to provide good quality real-time parameter estimation could potentially be important in models where time scales are much shorter than those pertaining to COVID-19. In such cases, the training time of PINNs to estimate parameters makes it less suitable for practical purposes, although we recognize that it may serve other useful purposes, such as dealing with parameter estimation when there is a low volume of data available, taking advantage of utilizing physics-based modeling to compensate for lack of available data. In applications where sparse and possibly noise data are available, these parameter estimation techniques often become less accurate. The PIR method was also explored for flow about a cylinder resulting in vortex shedding resulting in a K\'arm\'an vortex street. Using public data, it is demonstrated that with few sensors it is possible to estimate the dimensionless Reynolds number. This information is relevant to characterize the flow regime and provide estimates of the ratio of inertial forces to viscous forces in the fluid.

\section{Data Availability}

Upon acceptance of this paper, the data and codes used to produce the results presented in this manuscript will be made public in the GitHub repository linked below:
\begin{center}    \url{https://github.com/MEGAjosni/Physics-Informed-Regression}
\end{center}

A Julia software package containing all relevant PIR methods has been made publicly available:
\begin{center}          \url{https://github.com/MarcusGalea/PhysicsInformedRegression.jl} 
\end{center}


\printbibliography

@article{Colonius2008AConditions,
    title = {{A fast immersed boundary method using a nullspace approach and multi-domain far-field boundary conditions}},
    year = {2008},
    journal = {Computer Methods in Applied Mechanics and Engineering},
    author = {Colonius, Tim and Taira, Kunihiko},
    number = {25-28},
    month = {4},
    pages = {2131--2146},
    volume = {197},
    doi = {10.1016/J.CMA.2007.08.014},
    issn = {00457825}
}

@article{Psichogios1992AModeling,
    title = {{A hybrid neural network‐first principles approach to process modeling}},
    year = {1992},
    journal = {AIChE Journal},
    author = {Psichogios, Dimitris C. and Ungar, Lyle H.},
    number = {10},
    pages = {1499--1511},
    volume = {38},
    doi = {10.1002/AIC.690381003},
    issn = {15475905}
}

@article{Bock1984AProblems,
    title = {{A Multiple Shooting Algorithm for Direct Solution of Optimal Control Problems}},
    year = {1984},
    journal = {IFAC Proceedings Volumes},
    author = {Bock, H. G. and Plitt, K. J.},
    number = {2},
    month = {7},
    pages = {1603--1608},
    volume = {17},
    publisher = {Elsevier},
    url = {https://www.sciencedirect.com/science/article/pii/S1474667017612059},
    isbn = {0080316697},
    doi = {10.1016/S1474-6670(17)61205-9},
    issn = {1474-6670}
}

@article{Cooper2020ACommunities,
    title = {{A SIR model assumption for the spread of COVID-19 in different communities}},
    year = {2020},
    journal = {Chaos, Solitons and Fractals},
    author = {Cooper, Ian and Mondal, Argha and Antonopoulos, Chris G.},
    volume = {139},
    doi = {10.1016/j.chaos.2020.110057},
    issn = {09600779}
}

@article{Varah1982AEquations,
    title = {{A Spline Least Squares Method for Numerical Parameter Estimation in Differential Equations}},
    year = {1982},
    journal = {SIAM Journal on Scientific and Statistical Computing},
    author = {Varah, J. M.},
    number = {1},
    month = {3},
    pages = {28--46},
    volume = {3},
    publisher = {Society for Industrial {\&} Applied Mathematics (SIAM)},
    doi = {10.1137/0903003},
    issn = {0196-5204}
}

@article{Nguyen2023ACylinders,
    title = {{A state-of-the-art review of flows past confined circular cylinders}},
    year = {2023},
    journal = {Physics of Fluids},
    author = {Nguyen, Quang Duy and Lu, Wilson and Chan, Leon and Ooi, Andrew and Lei, Chengwang},
    number = {7},
    month = {7},
    volume = {35},
    publisher = {American Institute of Physics Inc.},
    doi = {10.1063/5.0157470},
    issn = {10897666}
}

@article{Chen2020APersons,
    title = {{A Time-dependent SIR model for COVID-19 with Undetectable Infected Persons}},
    year = {2020},
    journal = {IEEE Transactions on Network Science and Engineering},
    author = {Chen, Yi-Cheng and Lu, Ping-En and Chang, Cheng-Shang and Liu, Tzu-Hsuan},
    number = {4},
    month = {2},
    pages = {3279--3294},
    volume = {7},
    publisher = {IEEE Computer Society},
    url = {http://arxiv.org/abs/2003.00122},
    doi = {10.1109/TNSE.2020.3024723},
    arxivId = {2003.00122v6}
}

@article{Kingma2014Adam:Optimization,
    title = {{Adam: A Method for Stochastic Optimization}},
    year = {2014},
    journal = {3rd International Conference on Learning Representations, ICLR 2015 - Conference Track Proceedings},
    author = {Kingma, Diederik P. and Ba, Jimmy Lei},
    month = {12},
    publisher = {International Conference on Learning Representations, ICLR},
    url = {https://arxiv.org/abs/1412.6980v9},
    arxivId = {1412.6980}
}

@book{Isaacson1994AnalysisMathematics,
    title = {{Analysis of Numerical Methods (Dover Books on Mathematics)}},
    year = {1994},
    author = {Isaacson, Eugene and Keller, Herbert B},
    publisher = {Dover Publications},
    url = {http://www.amazon.com/exec/obidos/redirect?tag=citeulike07-20&path=ASIN/0486680290},
    isbn = {0486680290}
}

@article{Lagaris1997ArtificialEquations,
    title = {{Artificial Neural Networks for Solving Ordinary and Partial Differential Equations}},
    year = {1997},
    journal = {IEEE Transactions on Neural Networks},
    author = {Lagaris, I. E. and Likas, A. and Fotiadis, D. I.},
    number = {5},
    month = {5},
    pages = {987--1000},
    volume = {9},
    url = {http://arxiv.org/abs/physics/9705023},
    doi = {10.1109/72.712178},
    arxivId = {physics/9705023v1}
}

@article{GunesBaydin2015AutomaticSurvey,
    title = {{Automatic differentiation in machine learning: a survey}},
    year = {2015},
    journal = {Journal of Machine Learning Research},
    author = {G{\"{u}}ne{\c{s}} Baydin, Atılım and Pearlmutter, Barak A. and Andreyevich Radul, Alexey and Mark Siskind, Jeffrey},
    month = {2},
    pages = {1--43},
    volume = {18},
    publisher = {Microtome Publishing},
    url = {https://arxiv.org/abs/1502.05767v4},
    issn = {15337928},
    arxivId = {1502.05767}
}

@article{Moreno-Martin2024CollocationSystems,
    title = {{Collocation methods for second and higher order systems}},
    year = {2024},
    journal = {Autonomous Robots},
    author = {Moreno-Mart{\'{i}}n, Siro and Ros, Lluís and Celaya, Enric},
    number = {1},
    volume = {48},
    doi = {10.1007/s10514-023-10155-z},
    issn = {15737527}
}

@article{Shaier2022Data-DrivenNetworks,
    title = {{Data-Driven Approaches for Predicting Spread of Infectious Diseases Through DINNs: Disease Informed Neural Networks}},
    year = {2022},
    journal = {Letters in Biomathematics},
    author = {Shaier, Sagi and Raissi, Maziar and Seshaiyer, Padmanabhan},
    number = {1},
    month = {2},
    pages = {71--105},
    volume = {9},
    publisher = {Intercollegiate Biomathematics Alliance},
    doi = {10.30707/LIB9.1.1681913305.249476},
    issn = {23737867},
    arxivId = {2110.05445}
}

@article{Rudy2016Data-drivenEquations,
    title = {{Data-driven discovery of partial differential equations}},
    year = {2016},
    author = {Rudy, Samuel H and Brunton, Steven L and Proctor, Joshua L and Nathan Kutz, J},
    pages = {123},
    publisher = {APS},
    isbn = {1609.06401v1},
    arxivId = {1609.06401v1}
}

@article{Rudy2019Data-drivenEquations,
    title = {{Data-driven identification of parametric partial differential equations}},
    year = {2019},
    journal = {SIAM Journal on Applied Dynamical Systems},
    author = {Rudy, Samuel and Alla, Alessandro and Brunton, Steven L. and Kutz, J. Nathan},
    number = {2},
    volume = {18},
    doi = {10.1137/18M1191944},
    issn = {15360040}
}

@article{Brunton2019Data-DrivenControl,
    title = {{Data-Driven Science and Engineering: Machine Learning, Dynamical Systems, and Control}},
    year = {2019},
    journal = {Data-Driven Science and Engineering: Machine Learning, Dynamical Systems, and Control},
    author = {Brunton, Steven L. and Kutz, J. Nathan},
    month = {1},
    pages = {1--472},
    publisher = {Cambridge University Press},
    isbn = {9781108380690},
    doi = {10.1017/9781108380690}
}

@article{Raissi2018DeepEquations,
    title = {{Deep Hidden Physics Models: Deep Learning of Nonlinear Partial Differential Equations}},
    year = {2018},
    journal = {Journal of Machine Learning Research},
    author = {Raissi, Maziar},
    month = {1},
    pages = {1--24},
    volume = {19},
    publisher = {Microtome Publishing},
    url = {https://arxiv.org/pdf/1801.06637},
    issn = {15337928},
    arxivId = {1801.06637}
}

@article{Lecun2015DeepLearning,
    title = {{Deep learning}},
    year = {2015},
    journal = {Nature},
    author = {Lecun, Yann and Bengio, Yoshua and Hinton, Geoffrey},
    number = {7553},
    month = {5},
    pages = {436--444},
    volume = {521},
    publisher = {Nature Publishing Group},
    doi = {10.1038/NATURE14539},
    issn = {14764687},
    pmid = {26017442}
}

@article{Sirignano2018DGM:Equations,
    title = {{DGM: A deep learning algorithm for solving partial differential equations}},
    year = {2018},
    journal = {Journal of Computational Physics},
    author = {Sirignano, Justin and Spiliopoulos, Konstantinos},
    month = {12},
    pages = {1339--1364},
    volume = {375},
    publisher = {Academic Press Inc.},
    doi = {10.1016/J.JCP.2018.08.029},
    issn = {10902716},
    arxivId = {1708.07469}
}

@article{Bock2015DirectModels,
    title = {{Direct Multiple Shooting and Generalized Gauss-Newton Method for Parameter Estimation Problems in ODE Models}},
    year = {2015},
    author = {Bock, Hans Georg and Kostina, Ekaterina and Schl{\"{o}}der, Johannes P.},
    pages = {1--34},
    publisher = {Springer, Cham},
    url = {https://link.springer.com/chapter/10.1007/978-3-319-23321-5_1},
    isbn = {978-3-319-23321-5},
    doi = {10.1007/978-3-319-23321-5{\_}1},
    issn = {2191-3048}
}

@article{Brunton2016DiscoveringSystems,
    title = {{Discovering governing equations from data by sparse identification of nonlinear dynamical systems}},
    year = {2016},
    journal = {Proceedings of the National Academy of Sciences of the United States of America},
    author = {Brunton, Steven L. and Proctor, Joshua L. and Kutz, J. Nathan},
    number = {15},
    volume = {113},
    doi = {10.1073/pnas.1517384113},
    issn = {10916490}
}

@article{Metropolis1953EquationMachines,
    title = {{Equation of State Calculations by Fast Computing Machines}},
    year = {1953},
    journal = {The Journal of Chemical Physics},
    author = {Metropolis, Nicholas and Rosenbluth, Arianna W. and Rosenbluth, Marshall N. and Teller, Augusta H. and Teller, Edward},
    number = {6},
    month = {6},
    pages = {1087--1092},
    volume = {21},
    publisher = {AIP Publishing},
    url = {/aip/jcp/article/21/6/1087/202680/Equation-of-State-Calculations-by-Fast-Computing},
    doi = {10.1063/1.1699114},
    issn = {0021-9606}
}

@article{Baake1992FittingData,
    title = {{Fitting ordinary differential equations to chaotic data}},
    year = {1992},
    journal = {Physical Review A},
    author = {Baake, E. and Baake, M. and Bock, H. G. and Briggs, K. M.},
    number = {8},
    month = {4},
    pages = {5524--5529},
    volume = {45},
    url = {https://link.aps.org/doi/10.1103/PhysRevA.45.5524},
    doi = {10.1103/PhysRevA.45.5524},
    issn = {10502947}
}

@article{Ellner2002FittingMatching,
    title = {{Fitting Population Dynamic Models to Time-Series Data by Gradient Matching}},
    year = {2002},
    journal = {Ecology},
    author = {Ellner, Stephen P. and Seifu, Yodit and Smith, Robert H.},
    number = {8},
    month = {8},
    pages = {2256--2270},
    volume = {83},
    publisher = {Wiley},
    doi = {10.2307/3072057},
    issn = {00129658}
}

@misc{Rackauckas2025HowLifestyle,
    title = {{How chaotic is chaos? How some AI for Science / SciML papers are overstating accuracy claims - Stochastic Lifestyle}},
    year = {2025},
    booktitle = {https://www.stochasticlifestyle.com/how-chaotic-is-chaos-how-some-ai-for-science-sciml-papers-are-overstating-accuracy-claims/},
    author = {Rackauckas, Christopher},
    month = {5},
    url = {https://www.stochasticlifestyle.com/how-chaotic-is-chaos-how-some-ai-for-science-sciml-papers-are-overstating-accuracy-claims/}
}

@article{Rumelhart1986LearningErrors,
    title = {{Learning representations by back-propagating errors}},
    year = {1986},
    journal = {Nature},
    author = {Rumelhart, David E. and Hinton, Geoffrey E. and Williams, Ronald J.},
    number = {6088},
    volume = {323},
    doi = {10.1038/323533a0},
    issn = {00280836}
}

@article{Elerian2001LikelihoodDiffusions,
    title = {{Likelihood inference for discretely observed nonlinear diffusions}},
    year = {2001},
    journal = {Econometrica},
    author = {Elerian, Ola and Chib, Siddhartha and Shephard, Neil},
    number = {4},
    pages = {959--993},
    volume = {69},
    publisher = {Blackwell Publishing Ltd},
    doi = {10.1111/1468-0262.00226},
    issn = {00129682}
}

@article{Kato2020Likelihood-basedModel,
    title = {{Likelihood-based strategies for estimating unknown parameters and predicting missing data in the simultaneous autoregressive model}},
    year = {2020},
    journal = {Journal of Geographical Systems},
    author = {Kato, Takafumi},
    number = {1},
    month = {1},
    pages = {143--176},
    volume = {22},
    publisher = {Springer Science and Business Media Deutschland GmbH},
    url = {https://link.springer.com/article/10.1007/s10109-019-00316-z},
    doi = {10.1007/S10109-019-00316-Z/METRICS},
    issn = {14355949}
}

@article{Mac2021ModelingModels,
    title = {{Modeling the coronavirus disease 2019 pandemic: A comprehensive guide of infectious disease and decision-analytic models}},
    year = {2021},
    journal = {Journal of Clinical Epidemiology},
    author = {Mac, Stephen and Mishra, Sharmistha and Ximenes, Raphael and Barrett, Kali and Khan, Yasin A. and Naimark, David M.J. and Sander, Beate},
    month = {4},
    pages = {133--141},
    volume = {132},
    publisher = {Elsevier Inc.},
    doi = {10.1016/J.JCLINEPI.2020.12.002},
    issn = {18785921},
    pmid = {33301904}
}

@article{Hestenes1969MultiplierMethods,
    title = {{Multiplier and gradient methods}},
    year = {1969},
    journal = {Journal of Optimization Theory and Applications},
    author = {Hestenes, Magnus R.},
    number = {5},
    month = {11},
    pages = {303--320},
    volume = {4},
    publisher = {Kluwer Academic Publishers-Plenum Publishers},
    url = {https://link.springer.com/article/10.1007/BF00927673},
    doi = {10.1007/BF00927673/METRICS},
    issn = {00223239}
}

@article{Lee1990NeuralEquations,
    title = {{Neural algorithm for solving differential equations}},
    year = {1990},
    journal = {Journal of Computational Physics},
    author = {Lee, Hyuk and Kang, In Seok},
    number = {1},
    month = {11},
    pages = {110--131},
    volume = {91},
    publisher = {Academic Press},
    url = {https://www.sciencedirect.com/science/article/pii/002199919090007N},
    doi = {10.1016/0021-9991(90)90007-N},
    issn = {0021-9991}
}

@article{Chen2018NeuralEquations,
    title = {{Neural Ordinary Differential Equations}},
    year = {2018},
    journal = {NIPs},
    author = {Chen, Ricky T. Q. and Rubanova, Yulia and Bettencourt, Jesse and Duvenaud, David},
    number = {NeurIPS},
    month = {6},
    pages = {31--60},
    volume = {109},
    url = {https://arxiv.org/abs/1806.07366v5},
    issn = {20385757},
    arxivId = {1806.07366}
}

@article{Dissanayake1994NeuralnetworkbasedEquations,
    title = {{Neural‐network‐based approximations for solving partial differential equations}},
    year = {1994},
    journal = {Communications in Numerical Methods in Engineering},
    author = {Dissanayake, M. W.M.G. and Phan‐Thien, N.},
    number = {3},
    pages = {195--201},
    volume = {10},
    doi = {10.1002/CNM.1640100303},
    issn = {10990887}
}

@article{Bard1974NonlinearEstimation,
    title = {{Nonlinear parameter estimation}},
    year = {1974},
    journal = {Academic Press},
    author = {Bard, Yonathan}
}

@article{Cheney2020NumericalComputing,
    title = {{Numerical mathematics and computing}},
    year = {2020},
    author = {Cheney, E. W.. and Kincaid, David.},
    publisher = {Brooks/Cole},
    isbn = {9780357670842}
}

@book{Butcher2016NumericalEquations,
    title = {{Numerical Methods for Ordinary Differential Equations}},
    year = {2016},
    booktitle = {Numerical Methods for Ordinary Differential Equations},
    author = {Butcher, John Charles},
    edition = {3},
    month = {8},
    pages = {1--513},
    publisher = {John Wiley {\&} Sons},
    isbn = {9781119121534},
    doi = {10.1002/9781119121534}
}

@article{Lodhi2022NumericalMethod,
    title = {{Numerical solution of non-linear Bratu-type boundary value problems via quintic B-spline collocation method}},
    year = {2022},
    journal = {AIMS Mathematics 2022 4:7257},
    author = {Lodhi, Ram Kishun and Aldosary, Saud Fahad and Nisar, Kottakkaran Sooppy and Alsaadi, Ateq},
    number = {4},
    pages = {7257--7273},
    volume = {7},
    publisher = {American Institute of Mathematical Sciences},
    url = {http://www.aimspress.com/article/doi/10.3934/math.2022405},
    doi = {10.3934/MATH.2022405},
    issn = {2473-6988}
}

@article{Liu1989OnOptimization,
    title = {{On the limited memory BFGS method for large scale optimization}},
    year = {1989},
    journal = {Mathematical Programming},
    author = {Liu, Dong C. and Nocedal, Jorge},
    number = {1-3},
    month = {8},
    pages = {503--528},
    volume = {45},
    publisher = {Springer-Verlag},
    doi = {10.1007/BF01589116},
    issn = {00255610}
}

@article{Ramsay2007ParameterApproach,
    title = {{Parameter estimation for differential equations: A generalized smoothing approach}},
    year = {2007},
    journal = {Journal of the Royal Statistical Society. Series B: Statistical Methodology},
    author = {Ramsay, J. O. and Hooker, G. and Campbell, D. and Cao, J.},
    number = {5},
    month = {11},
    pages = {741--796},
    volume = {69},
    doi = {10.1111/J.1467-9868.2007.00610.X},
    issn = {13697412}
}

@article{Beck1977ParameterScience,
    title = {{Parameter estimation in engineering and science}},
    year = {1977},
    author = {Beck, J. V.. and Arnold, Kenneth J..},
    pages = {501},
    publisher = {Wiley},
    url = {https://books.google.com/books/about/Parameter_Estimation_in_Engineering_and.html?hl=da&id=-opRAAAAMAAJ},
    isbn = {0471061182}
}

@article{Peifer2007ParameterShooting,
    title = {{Parameter estimation in ordinary differential equations for biochemical processes using the method of multiple shooting}},
    year = {2007},
    journal = {IET Systems Biology},
    author = {Peifer, M. and Timmer, J.},
    number = {2},
    month = {3},
    pages = {78--88},
    volume = {1},
    doi = {10.1049/IET-SYB:20060067},
    issn = {17518849},
    pmid = {17441551}
}

@article{Hao2021ParameterApplication,
    title = {{Parameter Estimation of the Lotka-Volterra Model with Fractional Order Based on the Modulation Function and Its Application}},
    year = {2021},
    journal = {Mathematical Problems in Engineering},
    author = {Hao, Ying and Guo, Mingshun},
    pages = {1--7},
    volume = {2021},
    publisher = {Hindawi Limited},
    doi = {10.1155/2021/6645059},
    issn = {15635147}
}

@article{Karniadakis2021Physics-informedLearning,
    title = {{Physics-informed machine learning}},
    year = {2021},
    journal = {Nature Reviews Physics},
    author = {Karniadakis, George Em and Kevrekidis, Ioannis G. and Lu, Lu and Perdikaris, Paris and Wang, Sifan and Yang, Liu},
    number = {6},
    month = {6},
    pages = {422--440},
    volume = {3},
    publisher = {Springer Nature},
    doi = {10.1038/S42254-021-00314-5},
    issn = {25225820}
}

@article{Borrel-Jensen2023Physics-informedBoundaries,
    title = {{Physics-informed neural networks for one-dimensional sound field predictions with parameterized sources and impedance boundaries}},
    year = {2023},
    journal = {JASA Express Letters},
    author = {Borrel-Jensen, Nikolas and Engsig-Karup, Allan P. and Jeong, Cheol-Ho},
    number = {12},
    month = {8},
    volume = {1},
    publisher = {American Institute of Physics},
    url = {http://arxiv.org/abs/2109.11313 http://dx.doi.org/10.1121/10.0009057},
    doi = {10.1121/10.0009057},
    arxivId = {2109.11313v5}
}

@article{Vogiatzoglou2025Physics-informedSpreading,
    title = {{Physics-informed neural networks for parameter learning of wildfire spreading}},
    year = {2025},
    journal = {Computer Methods in Applied Mechanics and Engineering},
    author = {Vogiatzoglou, K. and Papadimitriou, C. and Bontozoglou, V. and Ampountolas, K.},
    month = {2},
    pages = {117545},
    volume = {434},
    publisher = {North-Holland},
    url = {https://www.sciencedirect.com/science/article/abs/pii/S0045782524007990},
    doi = {10.1016/J.CMA.2024.117545},
    issn = {0045-7825},
    arxivId = {2406.14591}
}

@article{Raissi2019Physics-informedEquations,
    title = {{Physics-informed neural networks: A deep learning framework for solving forward and inverse problems involving nonlinear partial differential equations}},
    year = {2019},
    journal = {Journal of Computational Physics},
    author = {Raissi, Maziar and Perdikaris, Paris and Karniadakis, George Em},
    volume = {378},
    doi = {10.1016/j.jcp.2018.10.045},
    issn = {10902716}
}

@article{Hoerl1970RidgeProblems,
    title = {{Ridge Regression: Biased Estimation for Nonorthogonal Problems}},
    year = {1970},
    journal = {Technometrics},
    author = {Hoerl, Arthur E. and Kennard, Robert W.},
    number = {1},
    volume = {12},
    doi = {10.1080/00401706.1970.10488634},
    issn = {15372723}
}

@article{Kaheman2020SINDy-PI:SINDy-PI,
    title = {{SINDy-PI: A robust algorithm for parallel implicit sparse identification of nonlinear dynamics: SINDy-PI}},
    year = {2020},
    journal = {Proceedings of the Royal Society A: Mathematical, Physical and Engineering Sciences},
    author = {Kaheman, Kadierdan and Kutz, J. Nathan and Brunton, Steven L.},
    number = {2242},
    volume = {476},
    doi = {10.1098/rspa.2020.0279},
    issn = {14712946}
}

@article{Heydari2019SoftAdapt:Functions,
    title = {{SoftAdapt: Techniques for Adaptive Loss Weighting of Neural Networks with Multi-Part Loss Functions}},
    year = {2019},
    author = {Heydari, A. Ali and Thompson, Craig A. and Mehmood, Asif},
    month = {12},
    url = {https://arxiv.org/pdf/1912.12355},
    arxivId = {1912.12355}
}

@article{Meade1994SolutionNetworks,
    title = {{Solution of nonlinear ordinary differential equations by feedforward neural networks}},
    year = {1994},
    journal = {Mathematical and Computer Modelling},
    author = {Meade, A. J. and Fernandez, A. A.},
    number = {9},
    month = {11},
    pages = {19--44},
    volume = {20},
    publisher = {Pergamon},
    url = {https://www.sciencedirect.com/science/article/pii/089571779400160X},
    doi = {10.1016/0895-7177(94)00160-X},
    issn = {0895-7177}
}

@phdthesis{Rudd2013SolvingNetworks,
    title = {{Solving Partial Differential Equations Using Artificial Neural Networks}},
    year = {2013},
    author = {Rudd, Keith},
    url = {https://hdl.handle.net/10161/8197},
    school = {Duke University}
}

@techreport{Mller2020TekniskModellerne,
    title = {{Teknisk gennemgang af modellerne}},
    year = {2020},
    author = {M{\o}ller, Camilla Holten and Skov, Robert Leo and Gr{\ae}sb{\o}ll, Kaare and Christiansen, Lasse Engbo and Lehmann, Sune and Thygesen, Uffe Høgsbro and Mielke, Adam and Kirkeby, Carsten Thure and Denwood, Matt and St{\ae}rk-{\O}stergaard, Jacob and Lange, Theis and Petersen, Troels Christian and Heltberg, Mathias and Martiny, Emil Schou and Andreasen, Viggo and Mortensen, Laust Hvas and Nauta, Maarten and Eriksen, Rasmus Skytte and Lyngse, Frederik Plesner and Bager, Peter Michael},
    month = {6},
    url = {https://files.ssi.dk/teknisk-gennemgang-af-modellerne-10062020},
    institution = {Statens Serum Institut},
    address = {Copenhagen}
}

@article{Taira2007TheApproach,
    title = {{The immersed boundary method: A projection approach}},
    year = {2007},
    journal = {Journal of Computational Physics},
    author = {Taira, Kunihiko and Colonius, Tim},
    number = {2},
    month = {8},
    pages = {2118--2137},
    volume = {225},
    publisher = {Academic Press Inc.},
    doi = {10.1016/J.JCP.2007.03.005},
    issn = {10902716}
}

@article{Meade1994TheNetworks,
    title = {{The numerical solution of linear ordinary differential equations by feedforward neural networks}},
    year = {1994},
    journal = {Mathematical and Computer Modelling},
    author = {Meade, A. J. and Fernandez, A. A.},
    number = {12},
    month = {6},
    pages = {1--25},
    volume = {19},
    publisher = {Pergamon},
    url = {https://www.sciencedirect.com/science/article/pii/0895717794900957},
    doi = {10.1016/0895-7177(94)90095-7},
    issn = {0895-7177}
}

@article{Kiers1997WeightedAlgorithms,
    title = {{Weighted least squares fitting using ordinary least squares algorithms}},
    year = {1997},
    journal = {Psychometrika},
    author = {Kiers, Henk A.L.},
    number = {2},
    volume = {62},
    doi = {10.1007/BF02295279},
    issn = {00333123}
}

@misc{WorldHealthOrganization2025WHODenmark,
    title = {{WHO COVID-19 dashboard, COVID-19 Cases, Denmark}},
    year = {2025},
    author = {{World Health Organization}},
    url = {https://data.who.int/dashboards/covid19/cases?m49=208}
}

@article{Transtrum2009WhyChallenging,
    title = {{Why are nonlinear fits so challenging?}},
    year = {2009},
    journal = {Physical Review Letters},
    author = {Transtrum, M. K. and Machta, B. B. and Sethna, J. P.},
    number = {6},
    month = {12},
    volume = {104},
    url = {http://arxiv.org/abs/0909.3884 http://dx.doi.org/10.1103/PhysRevLett.104.060201},
    doi = {10.1103/PhysRevLett.104.060201},
    arxivId = {0909.3884v2}
}


\section{Appendix}

\subsection{Note on Initial Conditions for PINN Parameter Estimation}\label{section: note on IC}
\begin{figure}
    \centering
    \subfigure[Initial value: $I_{1,0}=10^{-2}$]{%
        \includegraphics{pdf_figures/S3I3R_const_v1_I123.pdf}
        \hspace{1cm}
        \includegraphics{pdf_figures/S3I3R_const_v1_params.pdf}
        \label{fig:init_1}
    }
    \hspace{0.5cm}
    \subfigure[Initial value: $I_{1,0}=10^{-3}$]{%
        \includegraphics{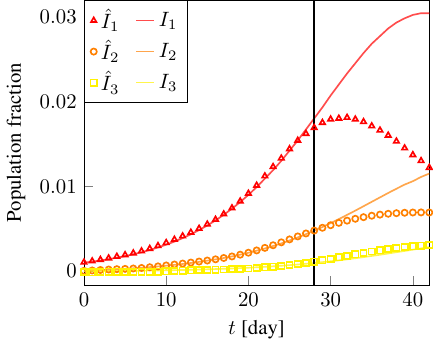}
        \hspace{1cm}
        \includegraphics{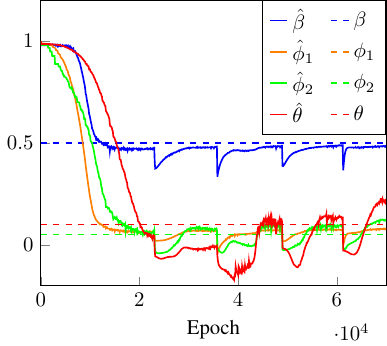}
        \label{fig:init_2}
    }
    \\
    \subfigure[Initial value: $I_{1,0}=10^{-4}$]{%
        \includegraphics{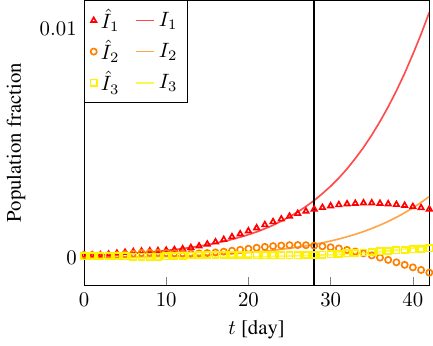}
        \hspace{1cm}
        \includegraphics{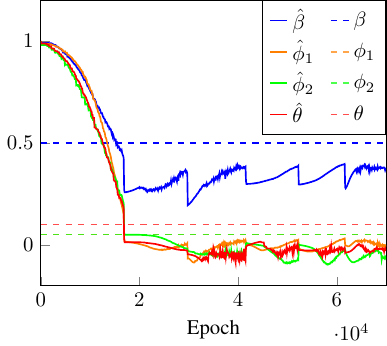}
        \label{fig:init_3}
    }
    \caption{PINN predictions (left) and parameter convergence (right) for various initial values of $I_1$.}
    \label{fig:init}
\end{figure}

As stated, the training data is generated using some initial conditions $X_0$ (and parameters). For simplicity, the models always assume that some small proportion of the population is infected and the rest is susceptible. This choice of proportion has an impact on the performance of the model and its ability to estimate the parameters. Figure \ref{fig:init} shows three models trained using $I_1=0.01,I_1=0.001$ and $I_1=0.0001$. 

In general, it seems as though the more dynamics in the period of interest, the better the estimates for the parameters are. This behavior was also seen with PIR, especially when sampling random parameters for the time-constant parameter experiments. Here, low values for $\beta$ would result in very small changes to the compartments, making it more difficult to accurately estimate the parameters.   

During the experiments, initial conditions were chosen such that the parameter convergence was not hindered by a lack of nonlinear dynamics in the period of interest.

\subsection{Note on Data Transformation} \label{sec: note on datatrans}
    A central problem when fitting SIR type models to real data is the discrepancy between how the relevant data is published, and how it is used in the model. Therefore, some transformation of the data presented in Table \ref{tab: data} is required. These transformations inherently introduce assumptions, which will be discussed in this section.
    
    First, access to the types of data presented in Table \ref{tab: data}. 
\begin{table}[H]
    \centering
    \begin{tabular}{c|l}
        \toprule
        \textbf{Symbol}  & \textbf{Description}\\
        \midrule
        $\Delta I^+$ & Number of people receiving their first positive test, at any given day.\\
        $\Delta R_3$ & Number of people registered dead, at any given day\\ 
        $\Delta R_2$ & Number of newly fully vaccinated people at any given day.\\ 
        $\Delta I_2$ & Number of hospitalizations at any given day. \\
        $\Delta I_3$ & Number of Intensive care hospitalizations (ICU) at any given day. \\
        \bottomrule
    \end{tabular}
    \caption{Necessary data}
    \label{tab: data}
\end{table}
In addition, fixed values must be assumed to be for $\gamma_1$, $\gamma_2$ and $\gamma_3$. They are therefore not estimated with either method, as this would be contradictory to the fixation. 

\subsubsection{Data transformation for the SIR model} \label{Data trans basic}
    In order to be able to estimate the parameters of the SIR model, it is necessary to construct a state matrix $n \times 3$, $\mat{X}$, using the supplied data,
    \begin{equation}
        \mat{X} =
        \begin{bmatrix}
            \mat{S} \ | \ \mat{I} \ | \ \mat{R}
        \end{bmatrix}
        =
        \begin{bmatrix}
            \mat{x}_0 \\
            \mat{x}_1 \\
            \vdots \\
            \mat{x}_n
        \end{bmatrix}.
    \end{equation}
    $\mat{S}$, $\mat{I}$ and $\mat{R}$ are the state vectors for the compartments, all lengths $n$ corresponding to the number of days. On any given day $t$, the state is given as $\mat{x}_t = \left[ S_t, I_t, R_t \right]$ corresponding to the number of susceptible, infected, and recovered people. $\Delta \mat{I}^+$, gathered from table \ref{tab: data}, is used to generate the data $\mat{X}$ for the basic SIR model as follows. 
    
    On a given day, $t$, the number of newly infected is denoted $\Delta I_t^+$. The SIR model assumes a time-constant population size
    \begin{equation} \label{eq: SIR_const_pop}
        N = S_t + I_t + R_t,
    \end{equation}
    and that each newly infected comes from the susceptible state. Formally, $\Delta S_t = -\Delta I_t^+ $. Initializing the entire population as susceptible, that is $S_0 = N$, allows for recursive generation of the entire data set $\mat{S}$, as follows,
    \begin{equation*}
        S_t =
        \begin{cases}
            N,                      &  t = 0, \\
            S_{t-1} + \Delta S_t,   &  1 \leq t \leq n.
        \end{cases}
    \end{equation*}
    In the SIR model, $\gamma$ denotes the rate at which an infected person recovers, thus the average duration of the disease length for a Covid-19 patient is $\frac{1}{\gamma}$ days. As the daily infected data are given as newly infected rather than the total number of infected, $\mat{I}$ must be estimated. Suppose, therefore that the total number of infected on any given day $t$, is the sum of newly infected over the last $\frac{1}{\gamma}$ days. Following this assumption, $\mat{I}$ is constructed as follows,
    \begin{equation*}
        I_t = 
        \begin{cases}
            \sum\limits_{k=0}^t\Delta I_k^+,       & 0 \leq t < \frac{1}{\gamma}, \\
            \sum\limits_{k=t-\frac{1}{\gamma}}^t \Delta I_k^+,    & \frac{1}{\gamma} \leq t \leq n.
       \end{cases}
    \end{equation*}
    Lastly, using equation \eqref{eq: SIR_const_pop}, $\mat{R}$ is computed as
    \begin{equation*}
        \mat{R} = N - (\mat{S} + \mat{I}).
    \end{equation*}

\subsubsection{Data transformation for the S3I3R model}
    For the S3I3R model, the methodology of data transformation presented in section \ref{Data trans basic} is also used to construct a state matrix,
    \begin{equation*}
        \mat{X} = 
        \begin{bmatrix}
            \mat{S} \ | \ \mat{I_1} \ | \ \mat{I_2} \ |\ \mat{I_3}\ |\ \mat{R_1}\ |\ \mat{R_2}\ |\ \mat{R_3}
        \end{bmatrix}
        =
        \begin{bmatrix}
            \mat{x}_0 \\
            \mat{x}_1 \\
            \vdots \\
            \mat{x}_n
        \end{bmatrix}.
    \end{equation*}
    Although similar, the infected compartment $\mat{I}$ of the SIR model differs from the counterpart of the S3I3R model, $\mat{I_1}$, as it now excludes hospitalized people. Therefore, it is necessary to subtract the newly hospitalized, denoted $\Delta I_{2_t}$, thus $\mat{I}_1$ is constructed as follows,
    \begin{equation*}
        I_{1_t} =
        \begin{cases}
            \sum\limits_{k=0}^t \left( \Delta I_{1_k}^+ \right) - \Delta I_{2_t}, & 0 \leq t < \frac{1}{\gamma_1}, \\
            \sum\limits_{k=t-\frac{1}{\gamma_1}}^t \Delta I_{1_k}^+ - \Delta I_{2_t}, & \frac{1}{\gamma_1} \leq t \leq n.
       \end{cases}
    \end{equation*}
    The data for $\mat{I}_2$ are constructed in a similar way. Let $\gamma_2$ be the hospitalized recovery rate, implying an average hospitalization time of $\frac{1}{\gamma_2}$. Assume that the total number of hospitalized people is the amount of newly admitted accumulated over the previous $\frac{1}{\gamma_2}$ days. Then $\mat{I}_2$ is constructed as follows,
    \begin{equation*}
        I_{2_t} =
        \begin{cases}
            \sum\limits_{k=0}^t\Delta I_{2_k}, & 0 \leq t < \frac{1}{\gamma_2}, \\
            \sum\limits_{k=t-\frac{1}{\gamma_2}}^t \Delta I_{2_k}, &  \frac{1}{\gamma_2} \leq t \leq n.
       \end{cases}
    \end{equation*}
    
    $\Delta \mat{I_3}$ can be calculated similarly to $\Delta \mat{I_1}$ and $\Delta \mat{I_2}$ using the fixed parameter $\gamma_3$.
    
    The data for vaccinated and dead people, referred to in Table \ref{tab: data}, is given as daily occurrences, thus the total number of people in the corresponding compartments, $\mat{R_2}$ and $\mat{R_3}$, is the accumulation of all the preceding days.
    \begin{equation*}
        R_{2_t} = \sum_{k=0}^t \Delta R_{2_t}, \quad 0 \leq t \leq n,
    \end{equation*}
    \begin{equation*}
        R_{3_t} = \sum_{k=0}^t \Delta R_{3_t}, \quad 0 \leq t \leq n.
    \end{equation*}
    
    The data for the susceptible compartment $\mat{S}$ also follows from the methodology used in \ref{Data trans basic}, although vaccinations are now needed. The number of people vaccinated from $\mat{S}$ is proportional to the fraction of $\mat{S}$ throughout the vaccinatable population. It is assumed that only compartments $\mat{S}$, $\mat{I_1}$ and $\mat{R_1}$ make up the vaccinatable population. 
    \begin{equation*}
        S_t = 
        \begin{cases}
            N, & t = 0, \\
            S_{t-1}-\Delta I_{1_t}^+ - \Delta R_{2_t} \cdot \frac{S_{t-1}}{S_{t-1} + I_{1_{t-1}}+R_{1_{t-1}}}, &  1 \leq t \leq n.
       \end{cases}
    \end{equation*}

    As the data for 6 out of 7 compartments are known, the last compartment $\mat{R_1}$ is calculated using the property that the S3I3R model assumes a constant population,
    \begin{equation*}
        \mat{R}_1 = N - (\mat{S} + \mat{I}_1 + \mat{I}_2 + \mat{I}_3 - \mat{R}_2 - \mat{R}_3).
    \end{equation*}

\subsection{Solution formula for regularized ordinary least squares (ridge regression)} \label{sec:regLSQproof}
\textit{
    If $\mat{A}{\mat\omega}=\mat{b}$ is an overdetermined linear system, then $\mat{\omega}^* = (\mat{A}^T \mat{A} + \lambda \mat{I})^{-1} \mat{A}^T \mat{b}$ is the solution to the ridge-regularized least squares problem,
}
\begin{equation*}
    \min_{\mat{\omega}} \|\mat{A\omega} - \mat{b}\|_2^2 + \lambda \|{\mat{\omega}}\|_2^2,
\end{equation*}
where $\lambda$ is a regularization parameter used to control overfitting or instability by penalizing large parameter values~\cite{Hoerl1970RidgeProblems}.

\textit{Proof}: Let \( \mat{A} \in \R^{m \times n} \) and \( \mat{b} \in \R^m \), with \( m \geq n \) and \( \text{rank}(\mat{A}) = n \). These assumptions imply that \( \mat{A}^T \mat{A} + \lambda \mat{I} \) is invertible for \( \lambda > 0 \). The regularized squared error between the left and right hand sides of the system \( \mat{A} {\mat{\omega}} = \mat{b} \) is
\begin{align*}
    E^2({\mat{\omega}}) = (\mat{A\omega} - \mat{b})^T (\mat{A\omega} - \mat{b}) + \lambda \mat{\omega}^T \mat{\omega}.
\end{align*}
Expanding the terms
\begin{align*}
\begin{aligned}
E^2(\mat{\omega}) &= (\mat{A\omega} - \mat{b})^T (\mat{A\omega} - \mat{b}) + \lambda \mat{\omega}^T \mat{\omega} \\
       &= \mat{\omega}^T \mat{A}^T \mat{A} \mat{\omega} - 2 \mat{b}^T \mat{A} \mat{\omega} + \mat{b}^T \mat{b} + \lambda \mat{\omega}^T \mat{\omega} \\
       &= \mat{\omega}^T (\mat{A}^T \mat{A} + \lambda \mat{I}) \mat{\omega} - 2 \mat{b}^T \mat{A} \mat{\omega} + \mat{b}^T \mat{b}.
\end{aligned}
\end{align*}

The derivative of \( E^2 \) with respect to \( \mat{\omega} \) is

\begin{align*}
\frac{\partial E^2}{\partial \mat{\omega}} = 2(\mat{A}^T \mat{A} + \lambda \mat{I})\mat{\omega} - 2\mat{A}^T \mat{b}.
\end{align*}

Since \( E^2(\mat{\omega}) \) is convex, \( \mat{\omega} \) is a minimizer if and only if the gradient is zero. Thus,

\begin{align*}
\frac{\partial E^2}{\partial \mat{\omega}} = 0 \quad \Leftrightarrow \quad (\mat{A}^T \mat{A} + \lambda \mat{I})\mat{\omega} = \mat{A}^T \mat{b} \quad \Leftrightarrow \quad \mat{\omega} = (\mat{A}^T \mat{A} + \lambda \mat{I})^{-1} \mat{A}^T \mat{b}. \quad \square
\end{align*}

\end{document}